\documentclass{article}
\usepackage{graphicx}
\usepackage[table,xcdraw]{xcolor}
\usepackage{ijcai22}

\usepackage{times}

\usepackage{soul}
\usepackage{url}
\usepackage[hidelinks]{hyperref}
\usepackage[utf8]{inputenc}
\usepackage[small]{caption}
\usepackage{graphicx}
\usepackage{amsmath}
\usepackage{booktabs}
\usepackage[ruled, linesnumbered]{algorithm2e}

\usepackage{times}
\usepackage{epsfig}

\usepackage{caption}
\usepackage{subcaption}

\usepackage{multirow}

\usepackage{enumitem,amssymb}
\newlist{todolist}{itemize}{2}
\setlist[todolist]{label=$\square$}
\usepackage{pifont}

\usepackage{catchfile}

\title{TOFU: \underline{T}owards \underline{O}bfuscated \underline{F}ederated \underline{U}pdates by Encoding Weight Updates into Gradients from Proxy Data}

\author{
Isha Garg\footnote{Authors contributed equally}\and
Manish Nagaraj\footnotemark[1]\And
Kaushik Roy\\
\affiliations
Purdue University\\
\emails
\{gargi, mnagara, kaushik\}@purdue.edu
}

\begin{document}

\maketitle
\let\oldsim\sim 
\renewcommand{\sim}{{\oldsim}}

\begin{abstract}
\vspace{-1.4mm}

Advances in Federated Learning and an abundance of user data have enabled rich collaborative learning between multiple clients, without sharing user data. This is done via a central server that aggregates learning in the form of weight updates. However, this comes at the cost of repeated expensive communication between the clients and the server, and concerns about compromised user privacy. The inversion of gradients into the data that generated them is termed data leakage. Encryption techniques can be used to counter this leakage, but at added expense. To address these challenges of communication efficiency and privacy, we propose TOFU, a novel algorithm which generates proxy data that encodes the weight updates for each client in its gradients. Instead of weight updates, this proxy data is now shared. Since input data is far lower in dimensional complexity than weights, this encoding allows us to send much lesser data per communication round. Additionally, the proxy data resembles noise, and even perfect reconstruction from data leakage attacks would invert the decoded gradients into unrecognizable noise, enhancing privacy. We show that TOFU enables learning with less than $1\%$ and $7\%$ accuracy drops on MNIST and on CIFAR-10 datasets, respectively. This drop can be recovered via a few rounds of expensive encrypted gradient exchange. This enables us to learn to near-full accuracy in a federated setup, while being $4\times$ and $6.6\times$ more communication efficient than the standard Federated Averaging algorithm on MNIST and CIFAR-10, respectively.

\end{abstract}

\vspace{-3.25mm}

\section{Introduction}
\label{sec:Introduction}
Federated learning is the regime in which many devices have access to localized data and communicate with each other either directly or through a central node to improve their learning abilities collaboratively, without sharing data. 
Here, we focus on the centralized setting, in which each device or `client’ learns on the data available to it and communicates the weight updates to a central node or `server', which aggregates the updates it receives from all the clients. The server propagates the aggregated update back to each client, thus enabling collaborative learning from data available to all devices, without actually sharing the data. The abundance of user data has enabled rich complex learning. However, this comes at the cost of increased computational or communication costs between the clients and the server, and with increasing concerns about compromised user privacy. Privacy of user data is a growing concern, and standard federated averaging techniques have been shown to be vulnerable to data leakage by inverting gradients into the data that generated them
\cite{dlg:zhu2020,ig:geiping2020}. Gradients can be encrypted to preserve privacy, but incurs further communication overhead. \cite{secureagg}. 


In this work, we focus on the communication between the clients and the server, a critical point for both communication and data leakage. Traditionally, in every communication round, each client shares its weight updates with the servers. To enable complex learning, the models are getting larger, growing to many millions of parameters \cite{vgg:simonyan2014,resnet:he2016}. To put things in context, a VGG13 model has $9.4$ million parameters, resulting in $36$ MB of data being shared per communication round, per device. Since each device only has limited data, the number of rounds needed for the server to reach convergence are orders of magnitude more than those needed by the individual clients, further heightening the communication cost and opportunities for privacy leaks. This cost quickly grows prohibitive in resource constrained settings with limited bandwidth. 

\begin{figure}[!t]
    \centering
    \includegraphics[width=1\columnwidth]{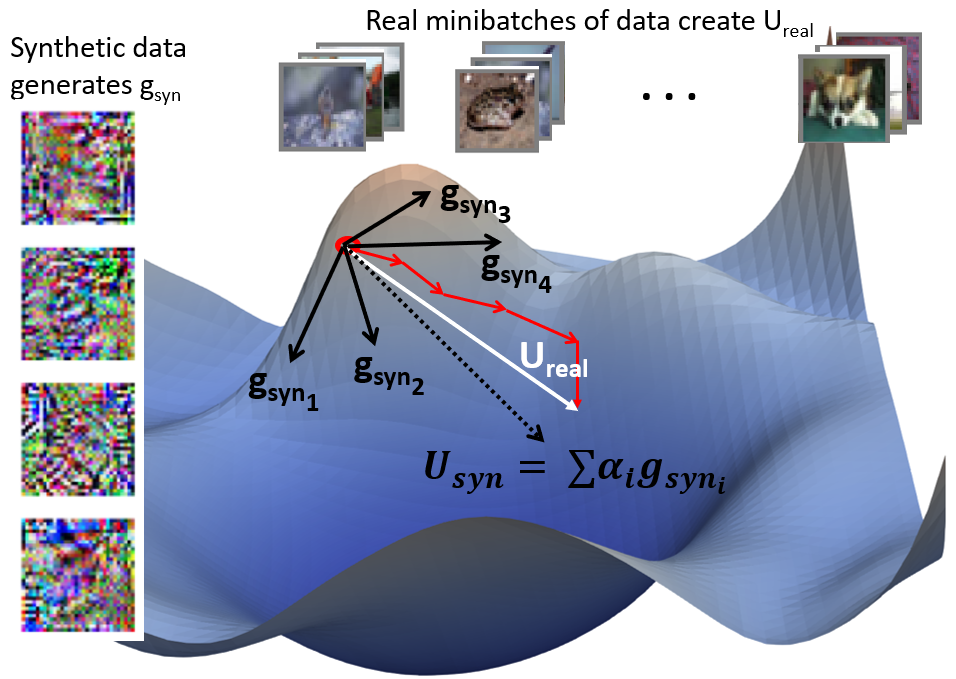}
    \vspace{-4mm}
    \caption{A pictorial representation of our encoding. The loss landscape shown in blue is taken from ~\protect\cite{landscape}, with the starting point marked with a red circle. Each client learns on some minibatches of real data, visualized on the top. The updates from these minibatches are shown with red arrows. The final weight update to be encoded and communicated, $U_{real}$, is calculated and shown in white. We construct a limited set of synthetic data that generates gradients $g_{syn}$ on the loss landscape, a weighted combination of which results in $U_{syn}$. The reconstruction algorithm optimizes these images and weights (denoted by $\alpha$) to maximize the cosine similarity between $U_{syn}$ and $U_{real}$. The synthetic images are visualized on the left and resemble noise, preserving user privacy.}
    \label{fig:overview}
    \vspace{-4mm}
\end{figure}

To address these concerns, we propose TOFU, a novel algorithm that works \underline{T}owards \underline{O}bfuscated \underline{F}ederated \underline{U}pdates, outlined pictorially in Figure \ref{fig:overview}. Here, each client generates synthetic proxy data whose combined gradient captures the weight update, and communicates this data instead of the weights. This mitigates two issues simultaneously. Data is much lower dimensional than gradients.
For context, CIFAR-10 images only have $3072$ pixels, and we show that TOFU needs under $100$ images to capture the weight updates well. Sending these images instead of the weight updates for VGG13 results in an order of magnitude reduction in communicated costs per round. Additionally, the synthetic data resembles noise (as shown in Figure \ref{fig:fake_images_attack}), and existent techniques would invert the gradients to this noisy data rather than the true data, thus enhancing privacy.

Since our method approximates gradients to reduce communication costs and enhance privacy, it results in a slight accuracy drop. Hence, we exchange proxy data for most of the early communication rounds, which are tolerant to noisy updates. Closer to convergence, updates are more precise and approximations are harmful. In these few communication rounds, we recover any accuracy drop by sharing the true full weight updates. In this phase, care needs to be taken to ensure privacy via expensive encryption techniques. Since this sensitive phase consists of far fewer communication rounds than the non-sensitive learning phase, any overheard resulting from this is mitigated by the communication efficiency achieved by sharing synthetic data instead of weight updates for most of the communication rounds. We show that we need only 3 and 15 full weight update rounds for MNIST and CIFAR-10, respectively, to recover any drop in accuracy.

This proposed hybrid approach provides both communication efficiency and privacy, without any loss in accuracy. We demonstrate TOFU on MNIST and CIFAR-10 datasets in both single device and federated setups. We show that learning with just synthetic proxy data shows less than $1\%$ and $7\%$ accuracy drops for MNIST and CIFAR-10, respectively. We extend this to a federated setup, with data distributed in an IID (Independent and Identically Distributed) fashion. We show that with a few additional rounds of full weight update, we can learn to accuracies comparable to FedAvg while achieving up to $4\times$ and $6.6\times$ better communication efficiency on MNIST and CIFAR-10, respectively. We emphasize that TOFU will results in increasing gains with increasing complexity of the models, since the size of weight updates will grow but the images stay a constant size. 

 

\section{Background}

\cite{fedavg:mcmahan2017} pioneered the field of Federated Learning and proposed the first algorithm for distributed private training, called FedAvg, which serves as our baseline. Here, only weight updates are shared with the server, which aggregates updates from all clients and shares them back with each client. A large part of the research focuses on the non-IID
setting, with varying data distributions allotted to clients. We focus on the IID setting in this work and direct readers to \cite{openprobs} for a better survey on non-IID methods. We focus on three key aspects in the IID setting: efficiency, privacy and accuracy, and discuss relevant works in each. 

\paragraph{Efficiency} Federated learning has two key areas of inefficiency: communication cost, both from client to server (up-communication) and from server to client (down-communication), and computational cost. The most potential for impact comes with decreasing client to server communication \cite{openprobs}. In our work, we target both up- and down-communication efficiency. Related works include quantization or sparsification of the weight updates \cite{strategies}, \cite{natcomp}, \cite{qsparse}, \cite{qsgd}. While they significantly improve communication efficiency, there have been concerns raised \cite{openprobs} about their compatibility with secure aggregation \cite{secureagg} and differential privacy techniques \cite{diffpriv}. Our method can be thought of as an indirect compression, by encoding updates into proxy inputs. Our proxy images are amenable to encryption, and can potentially be further quantized, resulting in additional savings. In this work, we focus on showing that it is possible to encode gradients into synthetic fake-looking data and still enable learning. Other methods restrict the structure of updates, such as to a low rank or a sparse matrix \cite{strategies}, or split the final network between the client and the server \cite{GKT}. We impose no constraints on learning, and focus on the standard case where each client has a synchronized model and equal accuracy on queries from any other client's dataset.

\paragraph{Privacy} Recent methods such as Inverting Gradients (IG) \cite{ig:geiping2020} and Deep Leakage from Gradients (DLG) \cite{dlg:zhu2020} have shown that gradients can be inverted into the training images that generated them, violating user data privacy. This is a significant concern, which we circumvent by showing that our proxy data looks like noise, and hence even perfect inversion by these techniques would only resemble noise. Other methods to secure gradients from attack involve encryption and differential privacy techniques that add additional computational expense \cite{secureagg}, \cite{diffpriv}. These methods are compatible with our proxy data, should the need for extra encryption arise. Additionally, encrypting our proxy data will be less costly since standard encryption costs are proportional to the size of the vector being encrypted \cite{secureagg}.

\paragraph{Accuracy}
Efforts to increase accuracy often focus on variance reduced SGD
\cite{scaffold}, \cite{yu2019linear} or on adaptive optimization and aggregation techniques \cite{DBLP:adaptfedopt}, \cite{mime}. Our method is orthogonal and compatible with such techniques.

\paragraph{Distillation} Recently, there has been increased interest in one-shot federated learning, wherein there is only one communication round. An approach that is conceptually similar to ours, called DOSFL  \cite{DOSFL:zhou2020} focuses on this setting. It is based on the dataset distillation \cite{datasetdistill} technique, in which the entire dataset is distilled into synthetic data. DOSFL uses this to distill the local data of each client and share that for one-shot learning. There are a few key differences between our method for synthetic data generation and dataset distillation. We generate proxy data that aligns its gradients to a desired weight update, whereas dataset distillation optimizes data for accuracy after learning on it. Dataset Distillation shows very large drops in accuracy for CIFAR-10 dataset ($\sim 26\%$) versus our single device results (Section 4.1), which shows under 8\% drop. DOSFL gets impressive results on MNIST, especially for a single round of communication but does not show results on larger datasets like CIFAR-10, presumably due to the significant drop in the baseline technique of dataset distillation.

\section{Methodology}
This section outlines the desired properties of the synthetic dataset, the algorithm to create it, and  the tradeoff between communication efficiency, privacy and accuracy.

\subsection{Desiderata of the Synthetic Dataset}

The generated synthetic dataset should have two properties - (a) it must be small in size in order to ensure communication efficiency and (b) it should not resemble the true data to ensure that data leakage attacks are unable to invert the proxy gradients into real data, thus ensuring privacy of the real data. We discuss these in more detail next. 

\textbf{Communication Efficiency:}
 Input data is much lower in dimensional complexity than gradients (for instance, $3072$ parameters per image in CIFAR-10 compared to $9.4$ million parameters for sending VGG13 weight updates). This allows us to attain the first goal of efficient communication. We experimentally show that $64$ images give us good results,
 which allows us to send $\sim 50\times$ lesser data per communication round.

\begin{figure}[!t]
\centering
\vspace{-1mm}
\subfloat[Synthetic images that encode weight updates]{%
  \includegraphics[clip,width=\columnwidth]{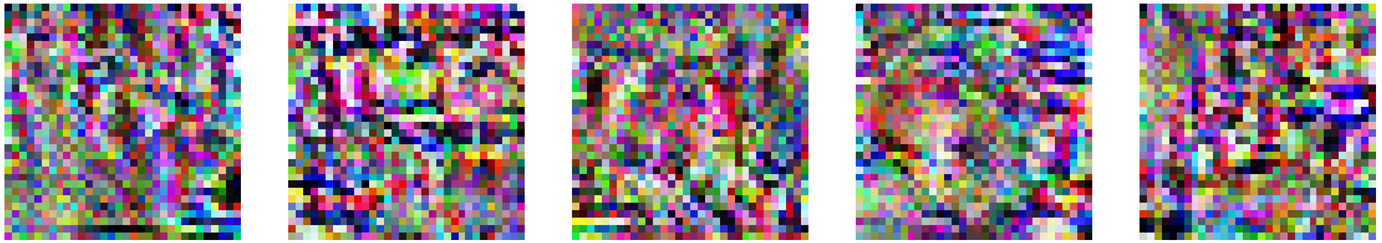}%
}

\subfloat[Images recovered by IG attack on synthetic images]{%
  \includegraphics[clip,width=\columnwidth]{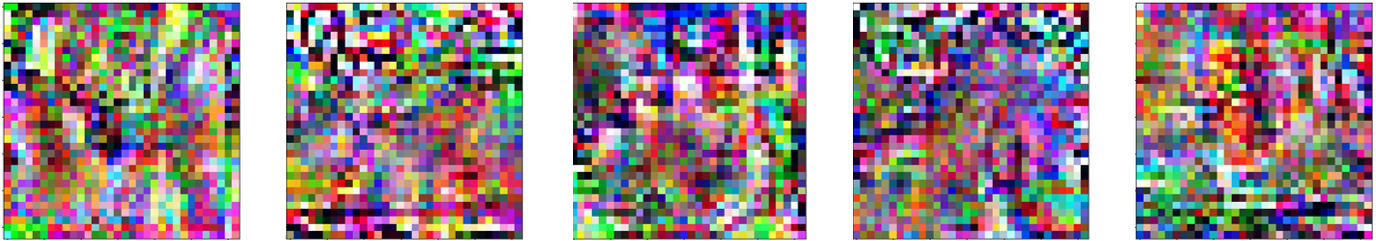}%
}
\vspace{-2mm}
\caption{The top row visualizes actual synthetic images ($x_{syn}$) generated by our algorithm. We show $5$ randomly picked images from a set of $32$ images encoding a weight update at the $200^{th}$ communication round of VGG13 on CIFAR-10, \textit{Synfreq} = $1$ epoch. The bottom row visualizes images recovered with the IG attack ~\protect\cite{ig:geiping2020} on the decoded weight update. Neither sets of images resemble CIFAR-10 images, protecting user data from leakage.}
\label{fig:fake_images_attack}
\vspace{-2mm}
\end{figure}

\textbf{Enhanced Privacy from Data Leakage:}
To attain the goal of privacy, we rely on the high dimensional, non-linear nature of neural networks to generate images that resemble noise to the human eye. We distill the weight update of a client after learning on many minibatches into a single minibatch. Combining the weight updates is not the same as combining inputs, and we observe that condensing the learning from many images into a smaller set results in images that visually do not conform to the true data distribution. The resulting gradient we generate is an approximation of the true weight update. This lossy compression buffers our gradient from data leakage as can be observed from the results of performing the IG attack (Inverting Gradients,  \cite{ig:geiping2020}) on our images, shown in Figure \ref{fig:fake_images_attack}. Even if IG attacks were to invert the images perfectly, the inverted images would still look like noise, circumventing data leakage. We employ some additional tricks to encourage obscurity of generated images. IG assumes availability of one-hot labels, or reconstructs one label per image. Instead, we use soft labels to further discourage reconstruction. Additionally, we weigh the gradients differently into the combined final gradient so that no true gradient is well represented. 

\textbf{Tradeoffs:}
The tradeoff cost associated here is two-fold: a) the clients have added computational complexity to create the synthetic data and b) the communication rounds needed for convergence increase since we introduce some error in the weight updates. We emphasize that our method is better applied to use-cases where the clients have computational resources but are limited in communication bandwidth or cost. Furthermore, communication efficiency has been identified as the major efficiency bottleneck, with the potential for most impact \cite{openprobs}. We hope that this raises more interest in creating efficient data-obfuscating methods. 
We account for the latter tradeoff of increased communication rounds when reporting our final efficiency ratios. 

\subsection{Creating the Synthetic Dataset}
We now detail the algorithm that distills the change in model parameters into a synthetic dataset during training. Taking inspiration from the IG attack, 
we optimize synthetic data to align the resulting gradient direction with the true weight update. To formalize, let this true weight update attained after a client learns on its true data be referred to as $U_{real}$. We want to generate a synthetic dataset, $$\mathcal{D}_{syn}=\{(x_{syn_i},y_{syn_i},\alpha_{syn_i}) ;\  \  i=1...N\}$$ where $N$ is the number of images in the synthetic dataset. $x_{syn_i}$ and $y_{syn_i}$ refer to the $i^{th}$ image and soft label respectively. The goal of reconstruction is that the combined gradient obtained upon forward and back-propagating all $\{x_{syn_i}, y_{syn_i}\} $ is aligned to the true weight update, $U_{real}$. 
Each synthetic datapoint generates a single gradient direction, and with $N$ datapoints in our synthetic dataset, we generate $N$ different gradient directions. Traditionally, if we were to treat these $N$ images as a minibatch, we would average the $N$ gradients. However, we take a weighted average of the gradients, allowing us to span a larger space. We jointly optimize these weights, referred to as $\alpha_{{syn}_i}; \sum_i \alpha_{{syn}_i} =1 $, along with the images and soft label. 

Next, we show that weighing the gradients of each image in the backward pass is the same as weighing the losses from each image on the forward pass, since derivative and summation can be interchanged. Formally, let $\theta$ be the weights of the model, and $\theta(x)$ the output of the model. Let the loss per synthetic datapoint, denoted by $L(\theta(x_{{syn}_i}),y_{{syn}_i} )$, be weighted by the respective $\alpha_{{syn}_i}$ and summed into $L_{syn}$, the overall loss of the synthetic dataset. Let the resulting gradient from backpropagating $L_{syn}$ be $U_{syn}$. Backpropagating $L_{syn}$ results in the desired weighted average of the individual gradients per datapoint, as shown:

\begin{align}
L_{syn} &= \sum_i \alpha_i L(\theta(x_{{syn}_i}),y_{{syn}_i}) \label{eq:spanned_loss} \\
U_{syn} &= \frac{\nabla L_{syn}}{\nabla \theta} = \sum_i \alpha_i \frac{\nabla L(\theta(x_{{syn}_i}),y_{{syn}_i})} {\nabla \theta} 
\end{align}

Standard cross entropy loss is used to calculate gradients from both the true and the synthetic data. The synthetic data is optimized by minimizing the reconstruction loss, $R_{loss}$, which is the cosine similarity between the true update $U_{real}$, and the synthetic update $U_{syn}$.  



\begin{equation}
R_{loss} = \bigg(1 - \dfrac{< U_{real}\cdot U_{syn} >}{\Vert U_{real} \Vert_{2} \Vert U_{syn} \Vert_2}\bigg)    
\label{eq:Similarity_metric}
\end{equation}

Since minimizing $R_{loss}$ only aligns the directions of gradients, we additionally send scaling values for each layer from $U_{real}$ to scale up $U_{syn}$. To avoid cluttering notation, we leave this out since it only adds extra parameters equal to the number of layers. In addition, we also use soft labels instead of the hard labels used in the real dataset. This provides more flexibility for the optimization algorithm to create a better alignment between synthetic and true updates, and discourages attacks like IG, which rely on one-hot labels. 
We use Adam \cite{Adam} to optimize the randomly initialized images to generate a gradient that aligns with the true weight updates. We use learning rates 0.1 for images, labels and spanning ratios, for 1000 iterations, decayed by a factor of 0.1 at the $375^{th}$, $625^{th}$, and the $875^{th}$ iteration. 


\subsection{TOFU: The Federated Learning Algorithm}
We now put all the parts together and describe how we utilize the synthesized dataset to enable communication efficient and private federated learning. All clients and the server have the same model initialization before the learning phase starts. Every client first trains on its private local data for a few minibatches and determines the true weight update, $U_{real}$, as the difference between the starting and ending weights. This true weight update is encoded using synthetic data ($\mathcal{D}_{syn}$) as described in Section 3.2. $\mathcal{D}_{syn}$ and then communicated in lieu of weight updates to the server. The server decodes the information by performing a single forward and backward pass to get the encoded weight update. The server repeats this for proxy data received from all clients and averages the decoded updates. To ensure efficiency during down-communication as well, the server encodes its own weight update due to aggregation into proxy data, and sends this back to all clients. The clients then update their local models by decoding this information. The process is repeated until convergence. This is summarized in Algorithm \ref{alg:CommEffFed} in Supplementary section \ref{appendix:algorithms}.  

\subsection{Efficiency-Privacy-Accuracy Tradeoff}

The weight update statistics change with accumulation over different number of batches and convergence progress. We now introduce the various hyperparameters that need to be tuned, and their effect on accuracy, privacy and efficiency. 

\textbf{Number of Synthetic Datapoints \textit{(Nimgs)}}: The size of the synthetic dataset transmitted per communication round has a direct impact on privacy, communication efficiency, and accuracy. While a larger synthetic dataset provides better accuracy as the encoding will be closer to the true weight update, the communication efficiency of the algorithm decreases since we have to communicate more data. We note that using 64-128 datapoints gives us the best empirical results. This also affects privacy since we will have larger approximation errors with smaller datasets, buffering us against attacks. 

\textbf{Synthesis Frequency \textit{(Synfreq)}:} This denotes how many minibatches of weight updates should be accumulated by the client before communicating with the server. In FedAvg, this is usually one epoch. We observe that having large \textit{Synfreq} results in larger accuracy drops, since a large accumulated weight cannot be well represented by few synthetic images. Larger \textit{Synfreq} also allows for enhanced privacy due to larger approximations. However, it degrades efficiency since we have to communicate more often per epoch. 

\textbf {Hybrid scheme of learning phases: \textit{(switch$_1$, switch$_2$)}}:  We divide learning into three phases, in decreasing order of error tolerance. The first phase is the early epochs of learning, where the gradient step is quite error tolerant since there is a strong direction of descent. We then switch to phase 2 at communication round \textit{switch$_1$}, where learning is more sensitive to the learning rate. In this phase, we scale $U_{syn}$ by (1-$R_{loss}$), capturing the cosine similarity between the true and the  synthetic update. This enables small steps to be taken if the synthetic data could not approximate the true update well. Switching too soon results in similar accuracy but with more communication rounds since we now take smaller steps for a larger duration. Switching too late leads to lower accuracy, presumably since we get stuck in a local optima. Once the learning saturates with this method, we briefly turn to phase 3 at communication round denoted by \textit{switch$_2$}, where we exchange full weight updates to regain any accuracy loss. To ensure privacy, we encourage expensive encryption of the weight updates here. Since it lasts for very few rounds (under 15), we do not sacrifice efficiency. We mark these switches in the learning curves shown in Supplementary section \ref{appendix:experiment_details}.
    




\section{Experimental Results and Discussion}
\label{sec:Results}
In this section, we first demonstrate TOFU on a single device setup to show that privacy preserved learning with only proxy data is possible. This setup can be thought of as a federated setup with only 1 client and no down-communication. We initialize two copies of the same network with the same weights. Network 1 represents the client, and learns on real data for \textit{Synfreq} number of batches, generating \textit{Nimgs} number of synthetic datapoints to send to Network 2, which emulates the server. Network 2 only learns on the synthetic data. Post communication, both the networks have the same weights since Network 1 knows how the synthetic data is going to update Network 2 and resets its own weights accordingly. For the single device experiments, we focus on getting the maximum accuracy from purely synthetic data, and hence we do not employ phase 3 of full weight update exchanges.

We then extend it to multiple clients in a federated setup. This has two encoding phases, the first carried out by each client to transmit their updates to the server (up-communication), and the second carried out by the server after aggregation from all clients (down-communication). Down-communication ensures that the weights of all clients and the server remain in sync after the end of each communication round. The experiments for Federated setup include phase 3 in order to circumvent any accuracy drop, and the emphasis in these experiments is on efficiency. All learning curves are shown in Supplementary section \ref{appendix:experiment_details}.

\subsection{Single Device Experiments}
\label{sec:single_device_results}

\begin{table}[t]
\centering
\setlength{\tabcolsep}{7pt}
\def\arraystretch{1.12}%
\begin{tabular}{|c|cc|cc|}
\hline
 & \multicolumn{2}{c|}{MNIST, LeNet5} & \multicolumn{2}{c|}{CIFAR-10, VGG13} \\ \hline
 & \begin{tabular}[c]{@{}c@{}}Max\\ Acc. (\%)\end{tabular} & \begin{tabular}[c]{@{}c@{}}Comm.\\ Eff.\end{tabular} & \begin{tabular}[c]{@{}c@{}}Max\\ Acc. (\%)\end{tabular} & \begin{tabular}[c]{@{}c@{}}Comm.\\ Eff.\end{tabular} \\ \hline
\rowcolor[HTML]{C0C0C0} 
Baseline & $99.32$ & $1.0\times$ & $90.42$ & $1.0\times$ \\ \hline
\textit{Nimgs} & \multicolumn{2}{c|}{\textit{Synfreq} = $100$} & \multicolumn{2}{c|}{\textit{Synfreq} = $200$} \\ \hline
$32$ & $97.59$ & $0.8\times$ & $79.48$ & $10.4\times$ \\
$64$ & $97.84$ & $0.9\times$ & $80.98$ & $7.1\times$ \\
$96$ & $97.93$ & $0.4\times$ & $82.32$ & $6.0\times$ \\
$128$ & $98.14$ & $0.3\times$ & $83.24$ & $5.2\times$ \\ \hline
\textit{Synfreq} & \multicolumn{2}{c|}{\textit{Nimgs} = $64$} & \multicolumn{2}{c|}{\textit{Nimgs} = $64$} \\ \hline
$50$ & $97.85$ & $0.3\times$ & $77.61$ & $7.0\times$ \\
$100$ & $97.84$ & $0.9\times$ & $82.44$ & $7.0\times$ \\
$200$ & $98.01$ & $0.9\times$ & $80.98$ & $7.1\times$ \\
$400$ & $97.21$ & $1.2\times$ & $72.58$ & $7.7\times$ \\ \hline
\end{tabular}
\caption{Single Device accuracies and efficiency ratios on MNIST and CIFAR-10 on synthetic data. The baseline model learnt on real data is shown in grey. Communication rounds required to reach maximum accuracy are shown in Supplementary section \ref{appendix:single_device}. }
\label{Table:Single_Device_compressed}
\end{table}
\paragraph{Setup} For MNIST, we use a slightly modified LeNet-5 network \cite{LeNet:lecun1998gradient}, with ReLU activation and maxpool. For CIFAR-10 we use a VGG-13 network \cite{vgg:simonyan2014}. For both the datasets, we use an SGD optimizer with learning rates 0.1 and 0.05 respectively. For CIFAR-10 and VGG13, we use a learning rate scheduler with a decay of 0.1 at the $80^{th}$ and $160^{th}$ epochs. We now discuss the results of tuning \textit{Synfreq} and \textit{Nimgs}. More details are shown in Supplementary section \ref{appendix:experiment_details}. The results for MNIST and CIFAR-10 are tabulated in Table \ref{Table:Single_Device_compressed}.

\paragraph{Varying \textit{Nimgs}} For both MNIST and CIFAR-10, Table \ref{Table:Single_Device_compressed} shows that increasing the synthetic dataset size improves accuracy, but reduces communication efficiency as we need to send more parameters per communication round. We achieve good accuracies by learning on only synthetic data, with only $\sim 1\%$ and $\sim 7\%$ drops for MNIST and CIFAR-10, respectively. Going further, we fix $\textit{Nimgs}=64$ for both CIFAR-10 and MNIST as a good tradeoff point. 

\paragraph{Varying \textit{Synfreq}} The trend for \textit{Synfreq} is lesser simple since the update statistics vary considerably as learning progresses. Empirically, we see that accumulating over too many or too few minibatches (\textit{Synfreq}$=400$ and $50$ respectively for CIFAR-10) results in large accuracy drops, as synthetic data fails to produce well aligned gradients. However, larger synthesis frequencies improve communication efficiency as we communicate lesser often, but increase the number of rounds required for convergence. Best results are obtained for \textit{Synfreq} of 100 for MNIST and 200 for CIFAR-10. 

\paragraph{Discussion} We successfully show that synthetic data can be used to learn, with $\sim 1\%$ and $\sim 7\%$ accuracy drops for MNIST and CIFAR-10, respectively, using only synthetic data. This drop is later recovered in the federated setup. There are two main observations. The first is that we do not show efficiency gains for MNIST in the single device setup. This is because we use LeNet5, which has only $\sim 60,000$ parameters and it is not particularly communication inefficient to send these many parameters. However, it is rare for networks today to be this small, and they are only growing larger, as evidenced by the results of Neural Architecture Search \cite{NAS}. The larger the networks, the more the savings that can be achieved with our method, since the size of updates will increase but the size of images remains constant. The second observation is that the number of communication rounds grow by $\sim 7\times$ for CIFAR-10 compared to the baseline model learnt on real data (convergence rounds required are shown in Table \ref{table:single_dev} in the Supplementary section \ref{appendix:single_device}). However, given that the data is $\sim 50\times$ smaller, this increase still allows for considerable efficiency of $7\times$, as shown in Table \ref{Table:Single_Device_compressed}.

\subsection{Federated Learning Experiments}
\label{sec:fed_learn_results}
\paragraph{Setup} We now discuss the federated experiments for 5 and 10 clients for CIFAR-10 and MNIST, respectively, with data distributed in an IID fashion. We compare with FedAvg as our baseline. We assume a participation rate of 1, since our aim is to show that communication-efficient federated learning is possible with obfuscated updates. The results are shown in Table \ref{Table:Federated_MNIST_compressed} for MNIST and Table \ref{Table:Federated_CIFAR_compressed} for CIFAR.

\begin{table}[!th]
\centering
\setlength{\tabcolsep}{7pt}
\def\arraystretch{1.12}%
\begin{tabular}{|ccccc|}
\hline
\multicolumn{5}{|c|}{MNIST, LeNet5, 10 Clients with IID distribution} \\ \hline
\multicolumn{1}{|c|}{} & \multicolumn{2}{c|}{\begin{tabular}[c]{@{}c@{}}Using Only \\ Synthetic Data\end{tabular}} & \multicolumn{2}{c|}{\begin{tabular}[c]{@{}c@{}}+ 3 Additional\\  Rounds of FedAvg\end{tabular}} \\ \hline
\multicolumn{1}{|c|}{} & \begin{tabular}[c]{@{}c@{}}Max.\\ Acc. (\%)\end{tabular} & \multicolumn{1}{c|}{\begin{tabular}[c]{@{}c@{}}Comm.\\ Eff.\end{tabular}} & \begin{tabular}[c]{@{}c@{}}Max\\ Acc. (\%)\end{tabular} & \begin{tabular}[c]{@{}c@{}}Comm.\\ Eff.\end{tabular} \\ \hline
\rowcolor[HTML]{C0C0C0} 
\multicolumn{1}{|c|}{\cellcolor[HTML]{C0C0C0}FEDAVG} & $98.91$ & \multicolumn{1}{c|}{\cellcolor[HTML]{C0C0C0}1.0$\times$} & - & - \\ \hline
\multicolumn{1}{|c|}{\textit{Nimgs}} & \multicolumn{4}{c|}{\textit{Synfreq} = 1 local epoch} \\ \hline
\multicolumn{1}{|c|}{32} & 95.39 & \multicolumn{1}{c|}{6.9$\times$} & 98.06 & 6.6$\times$ \\
\multicolumn{1}{|c|}{\textbf{64}} & \textbf{96.06} & \multicolumn{1}{c|}{\textbf{4.2$\times$}} & \textbf{98.14} & \textbf{4.1$\times$} \\
\multicolumn{1}{|c|}{96} & 96.77 & \multicolumn{1}{c|}{3.6$\times$} & 98.01 & 3.5$\times$ \\
\multicolumn{1}{|c|}{128} & 95.23 & \multicolumn{1}{c|}{4.2$\times$} & 98.07 & 4.1$\times$ \\ \hline
\multicolumn{1}{|c|}{\textit{Synfreq}} & \multicolumn{4}{c|}{\textit{Nimgs} = 64} \\ \hline
\multicolumn{1}{|c|}{25} & 92.00 & \multicolumn{1}{c|}{3.4$\times$} & 98.29 & 3.3$\times$ \\
\multicolumn{1}{|c|}{50} & 91.71 & \multicolumn{1}{c|}{7.9$\times$} & 98.19 & 7.5$\times$ \\
\multicolumn{1}{|c|}{\textbf{1 epoch}} & \textbf{96.06} & \multicolumn{1}{c|}{\textbf{4.2$\times$}} & \textbf{98.14} & \textbf{4.1$\times$} \\
\multicolumn{1}{|c|}{2 epochs} & 95.52 & \multicolumn{1}{c|}{3.1$\times$} & 97.91 & 3.1$\times$ \\ \hline
\end{tabular}
\caption{Accuracy and efficiency of the federated platform on MNIST. The baseline of FedAvg is shown in grey. }
\label{Table:Federated_MNIST_compressed}
\end{table}
\begin{table}[t]
\centering
\setlength{\tabcolsep}{7pt}
\def\arraystretch{1.12}%
\begin{tabular}{|ccccc|}
\hline
\multicolumn{5}{|c|}{CIFAR-10, VGG13, 5 Clients with IID distribution} \\ \hline
\multicolumn{1}{|c|}{} & \multicolumn{2}{c|}{\begin{tabular}[c]{@{}c@{}}Using Only \\ Synthetic Data\end{tabular}} & \multicolumn{2}{c|}{\begin{tabular}[c]{@{}c@{}}+ 15 Additional \\ Rounds of FedAvg\end{tabular}} \\ \hline
\multicolumn{1}{|c|}{} & \begin{tabular}[c]{@{}c@{}}Max.\\ Acc. (\%)\end{tabular} & \multicolumn{1}{c|}{\begin{tabular}[c]{@{}c@{}}Comm.\\ Eff.\end{tabular}} & \begin{tabular}[c]{@{}c@{}}Max\\ Acc. (\%)\end{tabular} & \begin{tabular}[c]{@{}c@{}}Comm.\\ Eff.\end{tabular} \\ \hline
\rowcolor[HTML]{C0C0C0} 
\multicolumn{1}{|c|}{\cellcolor[HTML]{C0C0C0}FEDAVG} & 88.73 & \multicolumn{1}{c|}{\cellcolor[HTML]{C0C0C0}1.0$\times$} & - & - \\ \hline
\multicolumn{1}{|c|}{\textit{Nimgs}} & \multicolumn{4}{c|}{\textit{Synfreq} = 1 local epoch} \\ \hline
\multicolumn{1}{|c|}{32} & 70.81 & \multicolumn{1}{c|}{26.7$\times$} & 88.31 & 7.8$\times$ \\
\multicolumn{1}{|c|}{\textbf{64}} & \textbf{72.31} & \multicolumn{1}{c|}{\textbf{16.6$\times$}} & \textbf{88.34} & \textbf{6.6$\times$} \\
\multicolumn{1}{|c|}{96} & 74.06 & \multicolumn{1}{c|}{10.0$\times$} & 88.48 & 5.3$\times$ \\
\multicolumn{1}{|c|}{128} & 74..03 & \multicolumn{1}{c|}{9.0$\times$} & 88.34 & 4.9$\times$ \\ \hline
\multicolumn{1}{|c|}{\textit{Synfreq}} & \multicolumn{4}{c|}{\textit{Nimgs} = 64} \\ \hline
\multicolumn{1}{|c|}{50} & 63.46 & \multicolumn{1}{c|}{19.1$\times$} & 88.36 & 7.0$\times$ \\
\multicolumn{1}{|c|}{100} & 68.21 & \multicolumn{1}{c|}{15.5$\times$} & 88.54 & 6.5$\times$ \\
\multicolumn{1}{|c|}{\textbf{1 epoch}} & \textbf{72.31} & \multicolumn{1}{c|}{\textbf{16.6$\times$}} & \textbf{88.34} & \textbf{6.6$\times$} \\ \hline
\end{tabular}
\caption{Accuracy and efficiency on  federated platform on CIFAR-10. The baseline of FedAvg is shown in grey. Results for additional 5 \& 10 rounds are in Table \ref{Table:Federated_CIFAR}, Supplementary section \ref{appendix:federated_cifar}.}
\label{Table:Federated_CIFAR_compressed}
\end{table}

\paragraph{Tuning \textit{Synfreq} and \textit{Nimgs}} For CIFAR-10 and MNIST, one local epoch for a client corresponds to 157 and 97 minibatches, respectively in this setup. We get the best results at a \textit{Synfreq} equal to one local epoch for both datasets, similar to FedAvg. Increasing dataset size (\textit{Nimgs}) results in better or comparable accuracy but sends more parameters per communication round. We get best accuracy for \textit{Nimgs} = $96$ for both datasets, seeing a drop of $2\%$ and $14\%$ for MNIST and CIFAR-10, with $4\times$ and $10\times$ communication efficiency, respectively. We note that the drop is roughly double what we see in single device, since each round incurs approximations from both up and down-communications. We show that we can recover this drop almost completely in just $3$ communication rounds for MNIST and $15$ rounds in CIFAR-10, still allowing for $3-6.5\times$ communication efficiency for MNIST and $5-8\times$ for CIFAR-10, as seen in the last columns of Tables \ref{Table:Federated_MNIST_compressed} and \ref{Table:Federated_CIFAR_compressed}. Additionally, we realize that showing maximum accuracy might skew communication efficiency in our benefit, since FedAvg reaches higher accuracy, which will naturally take more communication rounds. To account for this, we also report communication efficiency at iso-accuracy in Tables \ref{Table:IsoFederated_MNIST} and \ref{Table:IsoFederated_CIFAR} for MNIST and  CIFAR-10, respectively in Supplementary section \ref{appendix:experiment_details}. When we consider $72\%$ as the baseline accuracy for CIFAR-10, our efficiency of $5-8\times$ reduces to $1.5-2.5\times$. This shows that our method is still more efficient under stricter evaluation conditions.

\paragraph{Discussion} 
TOFU shares full weight updates for the last few epochs to regain full accuracy, which need to be encrypted to ensure privacy. For a conservative estimate while ensuring privacy, we assume that we need to encrypt all communication rounds for both methods, including synthetic data and full weight updates. Secure aggregation \cite{secureagg}, a commonly used protocol, shows that the communication cost is $\mathcal{O}(n+k)$ for the client and $\mathcal{O}(nk+n^2)$ for the server, where $k$ is the dimension of the vector being encrypted and $n$ is the number of clients. Comparing the encryption cost between FedAvg and TOFU for the same number of clients reduces to a ratio of the total parameters sent. This means that encryption retains the efficiency benefits of our method. The results show that TOFU can learn both MNIST and CIFAR-10, distributed in an IID setup, to near full accuracy with $\sim 4\times$ and $\sim 6.6\times$ communication efficiency (shown in bold in Tables \ref{Table:Federated_MNIST_compressed} and \ref{Table:Federated_CIFAR_compressed}), respectively.

\section{Conclusion}
 In the standard federated learning algorithm, clients carry out local learning on their private datasets for some minibatches, and communicate their weight updates to a central server. The central server aggregates the weight updates received from all clients, and communicates this update back to all clients. There are two major bottlenecks in this procedure; it is communication inefficient and it is shown that gradient and weight updates can be inverted into the data that generated them, violating user privacy. In this work, we introduce TOFU, a federated learning algorithm for communication efficiency and to enhance protection against data leakage via gradients. We encode the weight updates to be communicated into the gradients of a much smaller set of proxy data. The proxy data resembles noise and thus even perfect inversion from data leakage attacks will result in revealing this noise rather than user data. Additionally, data is far lower in dimensional complexity than gradients, improving communication efficiency. Since proxy data only approximates gradients, we observe a small drop in accuracy when learning only from this synthetic data. We show that the accuracy can be recovered by a very few communication rounds of full weight updates. To ensure privacy in this phase, we recommend encrypting the updates. Since these rounds are very few in comparison to the number of rounds where we exchange synthetic data, we are able to maintain communication efficiency. We show that we can learn the MNIST dataset, distributed between 10 clients and the CIFAR-10 dataset, distributed between 5 clients to accuracies comparable to FedAvg, with $\sim 4\times$ and $\sim 6.6\times$ communication efficiency, respectively. Availability of more data and compute capabilities has encouraged network sizes to grow. Since input data usually is of fixed dimensions, the communication efficiency advantages of TOFU are expected to grow with network size.

\vspace{-2mm}
\section{Acknowledgements}
This work was supported in part by the Center for Brain Inspired Computing (C-BRIC), one of the six centers in JUMP, a Semiconductor Research Corporation (SRC) program sponsored by DARPA, by the Semiconductor Research Corporation, the National Science Foundation, Intel Corporation, the DoD Vannevar Bush Fellowship, and by the U.S. Army Research Laboratory and the U.K.
Ministry of Defence under Agreement Number W911NF-16-3-0001.
\bibliographystyle{named}
\bibliography{references}

\clearpage

\newpage
\setcounter{secnumdepth}{3}

\section{Appendix}
\label{sec:appendix}


\subsection{TOFU: Pseudocode}
\label{appendix:algorithms}
Here, we provide the pseudocode for the operation of TOFU. 
Algorithm \ref{alg:Recon} describes the encoding of the weight update into gradients from synthetic data. The updates and gradients are treated as a tuple, one for each layer. The goal of the encoding algorithm is to produce a synthetic dataset $\mathcal{D}_{syn}$ which comprises of synthetic images ($x_{syn}$), synthetic soft labels ($y_{syn}$), spanning ratios ($\alpha_{syn}$), which along with the scaling ratio ($\gamma_{syn})$ encode the update $U_{syn}$. $\mathcal{D}_{syn}$ is optimized by minimizing the cosine similarity between this synthetically created weight update $U_{syn}$ and the true weight update $U_{real}$  as shown in Step \ref{alg:encode_optimization_line} of Algorithm \ref{alg:Recon}. Scaling ratios are then calculated to ensure the synthetic and target updates are of the same magnitude, shown in Step \ref{alg:encode_scaling_line} of Algorithm \ref{alg:Recon}.
\begin{algorithm}[htbp]
	\caption{Synthetic Dataset Creation (Encode)} \label{alg:Recon}
     \DontPrintSemicolon
	\SetKwInOut{Input}{Input}\SetKwInOut{Output}{Output}
	\Input {Target weight update: $U_{real}$, \\model weights: $\theta$, max iterations: $I_{max}$,
    	\\learning rates for optimizer: $\beta_x, \beta_y, \beta_\alpha$}
	\Output{Synthetic dataset: $D_{syn}$ in the form of images: $x_{syn}$, soft labels: $y_{syn}$, spanning ratios: $\alpha_{syn}$ and scaling ratios: $\gamma_{syn}$}
\begin{enumerate}
    \item Initialize $D_{syn}=[x_{syn},y_{syn},\alpha_{syn}] \sim \mathbb{N}(0,I)$\;

  \item For $i = 1$ to $I_{max}:$
  \label{alg:encode_optimization_line}
  \begin{enumerate}
    \item{Forward pass $D_{syn}$ and compute gradients}
    $L_{syn} = \sum_i \alpha_i L_{CrossEntropy}(\theta(x_{{syn}_i}),y_{{syn}_i})$ \\
    $U_{syn} = \frac{\nabla L_{syn}}{\nabla \theta}$ 
    
    \item  Optimize $D_{syn}$ to minimize cosine similarity between target and synthetic update \\
  $R_{loss}$ = \textit{cos\_sim}($U_{syn}, U_{real}) $ \;
  $x_{syn} \gets x_{syn} - \beta_x * \nabla_{x_{syn}} R_{loss}$ \;
  $y_{syn} \gets y_{syn} - \beta_y * \nabla_{y_{syn}} R_{loss}$\;
  $\alpha_{syn} \gets \alpha_{syn} - \beta_\alpha * \nabla_{\alpha_{syn}} R_{loss}$\;
   \end{enumerate}
 
 \item Compute the scaling ratio for each layer of the model:
 \label{alg:encode_scaling_line}
 For $l = 1$ to Number of layers:
 \begin{enumerate}
     \item  $\gamma_{syn,l} =\dfrac{\Vert U_{real,l}\Vert_2}{\Vert U_{syn,l}\Vert_2}$\;
 \end{enumerate}
  \end{enumerate}
\end{algorithm}
\begin{algorithm}[!h]
	\caption{Recreating Weight Update(Decode)} \label{alg:Decode}
     \DontPrintSemicolon
	\SetKwInOut{Input}{Input}\SetKwInOut{Output}{Output}
	\Input {Model weights: $\theta$, Synthetic dataset: $D_{syn}$,  in the form of images: $x_{syn}$, \\ soft labels: $y_{syn}$, spanning ratios: $\alpha_{syn}$ \\ and scaling ratios: $\gamma_{syn}$}
	\Output{Synthetic Weight Update: $U_{syn}$}
\begin{enumerate}
    \item{Forward pass $D_{syn}$ and compute gradients}
    $L_{syn} = \sum_i \alpha_i L_{CrossEntropy}(\theta(x_{{syn}_i}),y_{{syn}_i})$ \\
    $U' = \frac{\nabla L_{syn}}{\nabla \theta}$ 
    \item Scale the gradient layerwise with the scaling ratio
    \\ For $l = 1$ to Number of layers:
    \begin{enumerate}
        \item $U_{syn,l} = U'_{l} * \gamma_{syn,l}$
    \end{enumerate}
   \end{enumerate}
\end{algorithm}

Algorithm \ref{alg:Decode} describes how to decode the synthetically created dataset back into the update $U_{syn}$, which is just the reverse process of the Encode algorithm. The synthetic loss $L_{syn}$ is calculated by taking a weighted average of the cross entropy loss of all the synthetic images and labels. The gradient of this loss with respect to the model parameters, scaled according to the scaling ratios creates $U_{syn}$.
\begin{algorithm}[!h]
	\caption{TOFU, the Federated Setup} \label{alg:CommEffFed}
     \DontPrintSemicolon
\begin{enumerate}
    \item  Server Initializes Global model $\theta_{glob}$\;
 	\item For round$= 1, \dots ,R_{max}$
 	\begin {enumerate}
 	\item up-communication: For clients= 1, \dots ,k
 	\begin{enumerate}

 	\item Load $\theta_{loc} = \theta_{glob}$\;
	    \item 	While batch\_id $<$ synfreq: \ 

	     Each client learns on local data, with \\
	    $\theta_{new} =$ final weight
	    \item The target weight update is the difference between the starting and the final weight,  \
	    $U_{real} = \theta_{loc} - \theta_{new}$
	    \item Reset client weights to global weights \ 
	    $\theta_{new}  =  \theta_{glob}$

		\item  	Encode the true gradient into synthetic data and send to server:
 	$\mathcal{D}_{syn} = \textit{Encode}(U_{real}, \theta_{loc}, \vec{\beta})$ \;
 	
 	\item The server decodes the synthetic gradient 
    $U_{{syn}_k} = \textit{Decode}(\theta_{glob},\mathcal{D}_{syn})$.
    
    \end{enumerate}
    
    \item The server aggregates synthetic updates from all clients and decides target update for down-communication: \\
 	$U_{serv} =  \frac{\sum_k (U_{{syn}_k})}{k}$
 	
    \item Down-communication: server generates synthetic data to send the updated weights to all clients as well as updates its global model \\
 	$\mathcal{D}_{syn_{server}} = \textit{Encode}(U_{serv}, \theta_{glob}, \vec{\beta})$\;
 	$\theta_{glob} \gets \theta_{glob} - \textit{Decode}(\theta_{glob},\mathcal{D}_{syn_{server}})$ 
 	\item Each client decodes the synthetic data and updates the synced weights:\\
 	For clients= 1, \dots ,k:  \\
 	\begin{enumerate}
 	
 	\item $U_{{syn_{down,k}}} = \textit{Decode}(\theta_{glob},\mathcal{D}_{syn_{server}})$.
 	\item $\theta_{glob} \gets \theta_{glob} - (U_{syn_{down,k}})$ 

    \end{enumerate}
    \end{enumerate}
 	\end{enumerate}

\end{algorithm}

Next, we describe the application of TOFU in the federated setup in Algorithm \ref{alg:CommEffFed}. For each communication round, there are four phases as shown in the algorithm. The first phase is the up-communication. The clients first load their last seen global model as the local model. They each train on their local dataset to produce the target weight updated $U_{real}$. They encode this using Algorithm \ref{alg:Recon} and up-communicate this dataset to the server. The server decodes each of the clients synthetic dataset using Algorithm \ref{alg:Decode} to produce $U_{syn_{k}}$. 
The second phase is similar to that of standard federated learning where the synthetic updates from all clients are aggregated (averaged in our case). In the third, down-communication phase, the server encodes the average weight update into a synthetic dataset using a similar process described in the up-communication phase. The server also updates its own model based on this synthetic dataset to ensure all entities, clients and server, maintain the same global model at all communication rounds. The last phase, the client update phase, involves all the clients decoding the server's encoded weight update to update their global model.

\begin{figure}[!b]
\centering
\vspace{-1mm}
\subfloat[]{%
  \includegraphics[clip,width=\columnwidth]{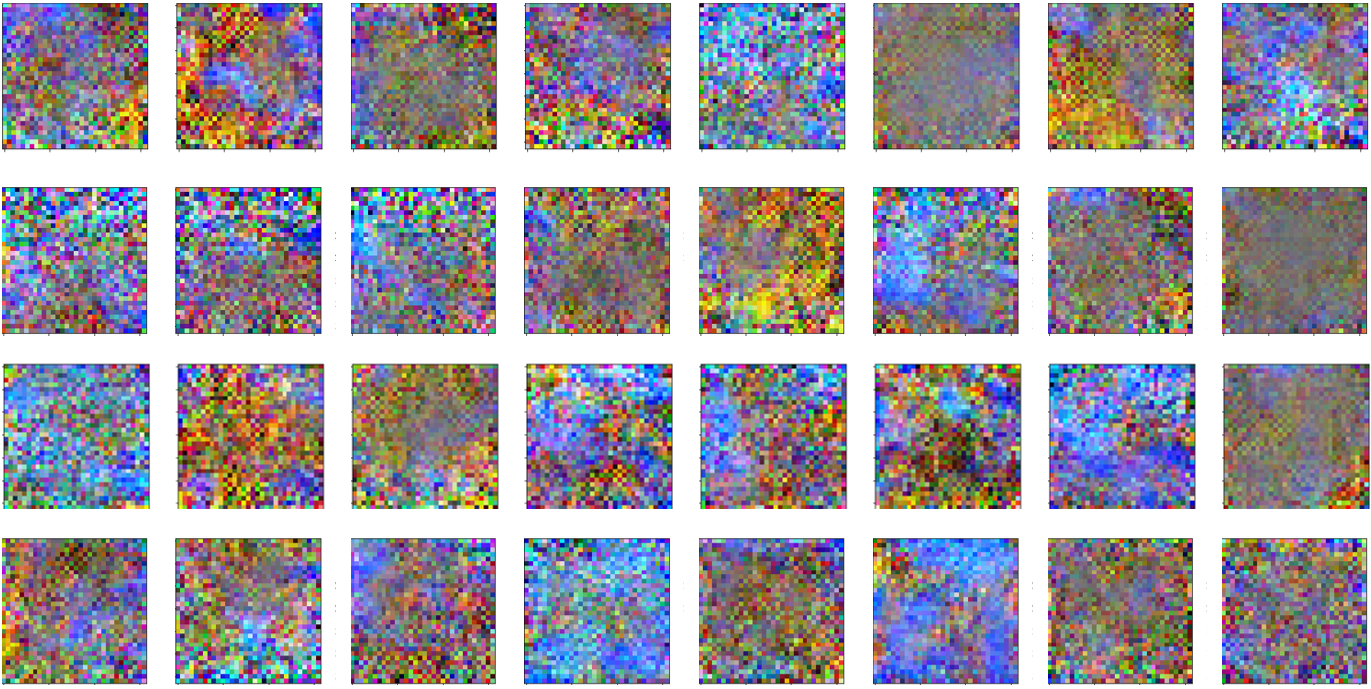}%
}

\subfloat[]{%
  \includegraphics[clip,width=\columnwidth]{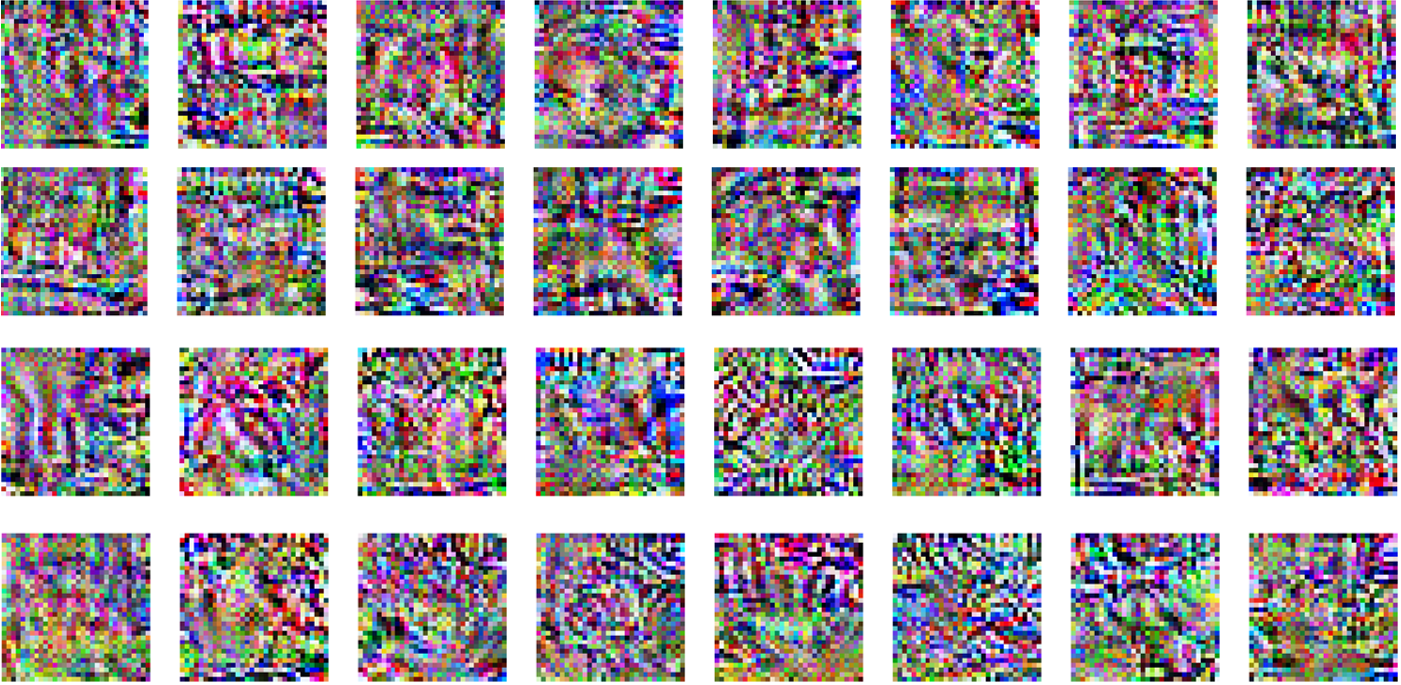}%
}

\subfloat[]{%
  \includegraphics[clip,width=\columnwidth]{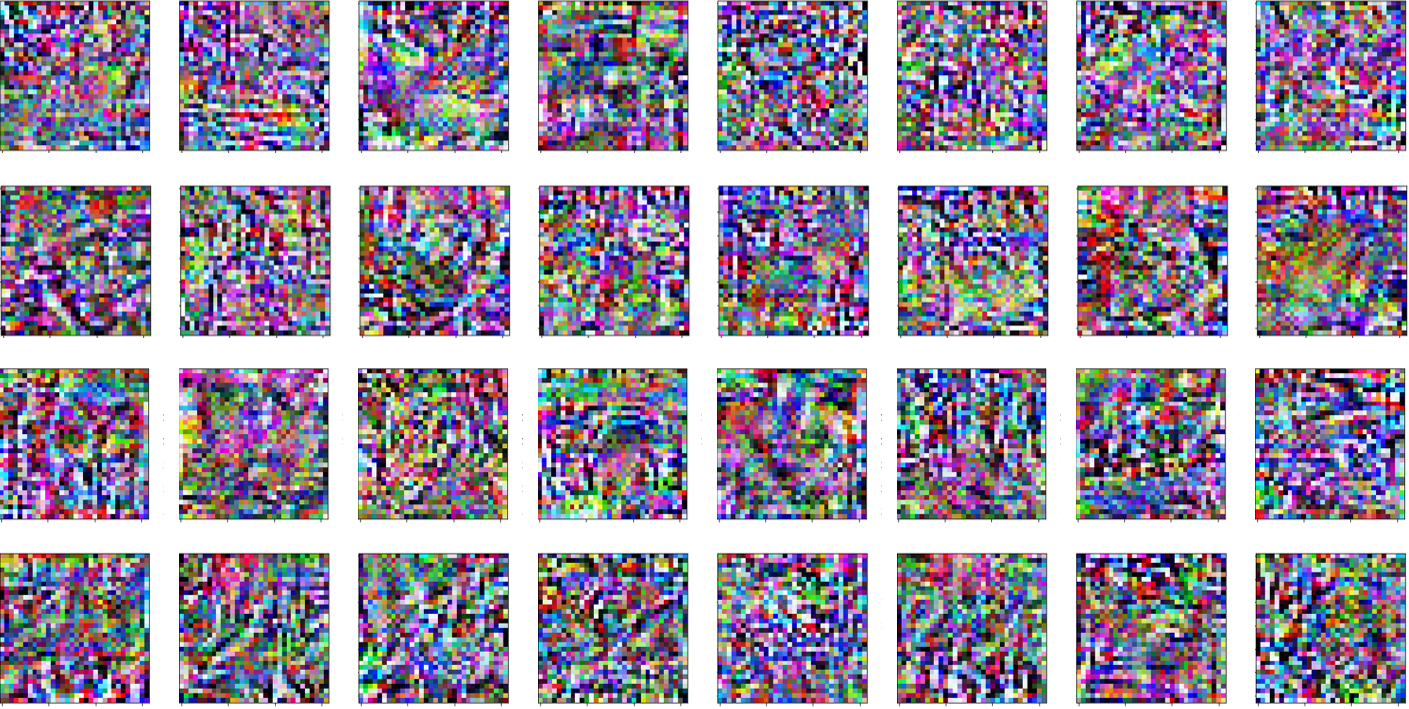}%
}
\vspace{-2mm}
\caption{Synthetic Images produced through the various communication rounds of TOFU. The rows, from top to bottom are, the 32 images encoding the up-communication of Client 1 at the start of the algorithm, at the $200^{th}$ communication round, and at the communication round with maximum accuracy ($593^{rd}$ communication round). This was performed on the CIFAR-10 dataset, distributed across 5 clients in an IID fashion, encoding at the end of every epoch. We see that the images do not bear any resemblance to the dataset.}
\label{fig:synthetic_images_appendix}
\vspace{-2mm}
\end{figure}

\subsection{Synthetic Images}
\label{appendix:Synthetic_images}
In this section, we aim to show how that the synthetic images do not resemble the dataset throughout the learning progress. We show results for experiments on CIFAR-10, with data distributed in an IID fashion across 5 clients. The frequency of synthesis is set to 1 local epoch and 32 images are used for both up and down-communication encoding. We have shown the encoded up-communication of Client 1 (arbitrarily chosen) when untrained (at the first communication), partially trained ($200^{th}$ communication round) and at max accuracy (70.81\% at the $593^{rd}$ communication round) in Figure \ref{fig:synthetic_images_appendix}. All data resembles noise, enhancing our protection against data leakage attacks.

\subsection{Experiment Details and Discussion}
\label{appendix:experiment_details}
In this section, we describe the details of the learning mechanisms we use in our experiment. We then discuss additional results that supplement the tables in the main text.

\subsubsection{Setup Details}
We use the same setup for both single device and federated experiments, and for both baselines and our experiments. For
MNIST, we use a slightly modified LeNet-5. We use relu non linearity and maxpool. There are two parallel instantiation of the second convolutional layer (with 16 filters) with their outputs summed together. We use an SGD optimizer and a learning rate 0.1 for 200 epochs. For CIFAR-10, we use a VGG13 with an SGD optimizer and a learning rate of 0.05, with a decay of $0.1$ first at 81 and then at 160 epochs for 200 epochs. All experiments were performed with a batchsize of 64. The baselines for the single device performance are standard learning on real data, and for federated is the FedAvg algorithm. For the encoding algorithms, we use an Adam optimizer with a learning rate of 0.1 set for images, labels and $\alpha$s, run for 1000 iterations.  We will release our code post publication for anonymization reasons.

\subsubsection{Single Device Performance}
\begin{table}[!bh]
\centering

\def\arraystretch{1.25}%
\setlength{\tabcolsep}{4.2pt}

\begin{tabular}{|c|cc|cc|}
\hline
 & \multicolumn{2}{c|}{MNIST} & \multicolumn{2}{c|}{CIFAR-10} \\ \hline
 & \multicolumn{1}{c|}{\begin{tabular}[c]{@{}c@{}}Max.\\ Acc(\%)\end{tabular}} & \begin{tabular}[c]{@{}c@{}}Comm. \\ Rnds (Eff.)\end{tabular} & \multicolumn{1}{c|}{\begin{tabular}[c]{@{}c@{}}Max.\\ Acc (\%)\end{tabular}} & \begin{tabular}[c]{@{}c@{}}Comm\\ Rnds (Eff)\end{tabular} \\ \hline
\rowcolor[HTML]{C0C0C0} 
Baseline & 99.32 & 495 (1.0$\times$) & 90.42 & 116 (1.0$\times$) \\ \hline
Nimgs & \multicolumn{2}{c|}{Synfreq = 100} & \multicolumn{2}{c|}{Synfreq = 200} \\ \hline
32 & 97.59 & 1194 (0.8$\times$) & 79.48 & 1057 (10.4$\times$) \\
64 & 97.84 & 512 (0.9$\times$) & 80.98 & 783 (7.1$\times$) \\
96 & 97.93 & 853 (0.4$\times$) & 82.32 & 613 (6.0$\times$) \\
128 & 98.14 & 890 (0.3$\times$) & 83.24 & 529 (5.2$\times$) \\ \hline
Synfreq & \multicolumn{2}{c|}{Nimgs = 64} & \multicolumn{2}{c|}{Nimgs = 64} \\ \hline
50 & 97.85 & 1587 (0.3$\times$) & 77.61 & 787 (7.0$\times$) \\
100 & 97.84 & 512 (0.9$\times$) & 82.44 & 786 (7.0$\times$) \\
200 & 98.01 & 540 (0.9$\times$) & 80.98 & 783 (7.1$\times$) \\
400 & 97.21 & 413 (1.2$\times$) & 72.58 & 722 (7.7$\times$) \\ \hline
\end{tabular}
\caption{Table showing Single Deivce accuracies on MNIST and CIFAR. The baseline model learnt on real data is shown in grey. We show communication rounds with corresponding communication efficiency compared to the baseline in brackets.}
\label{table:single_dev}
\end{table}

\label{appendix:single_device}
To observe how well our synthetic images capture the information of the dataset, we performed experiments on a single device setup. The baseline was regular learning on the true dataset. We varied the number of images and the frequency of synthesis as described in Section \ref{sec:single_device_results}.  A more detailed table of the results, showing the communication rounds needed till convergence is shown in Table \ref{table:single_dev}.

\begin{figure}[!h]
     \centering
     \begin{subfigure}[h]{\columnwidth}
         \centering
         \includegraphics[width=\textwidth]{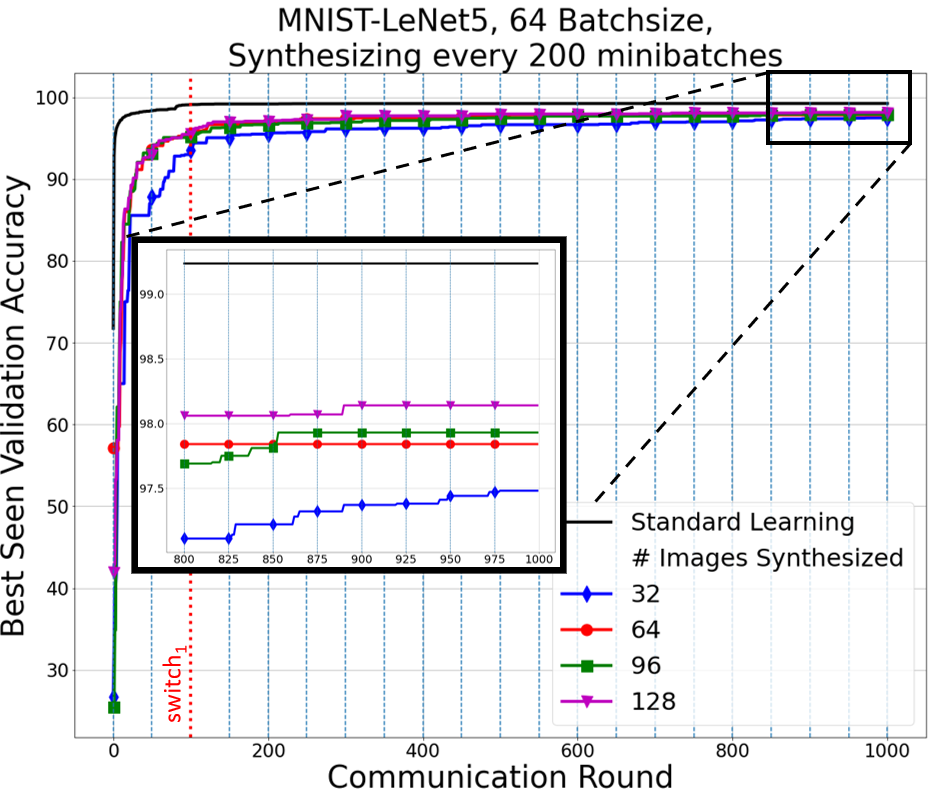}
         \caption{Variation of number of synthetic images.}
         \label{fig:MNIST_single_images}
     \end{subfigure}
     \hfill
     \begin{subfigure}[h]{\columnwidth}
         \centering
         \includegraphics[width=\textwidth]{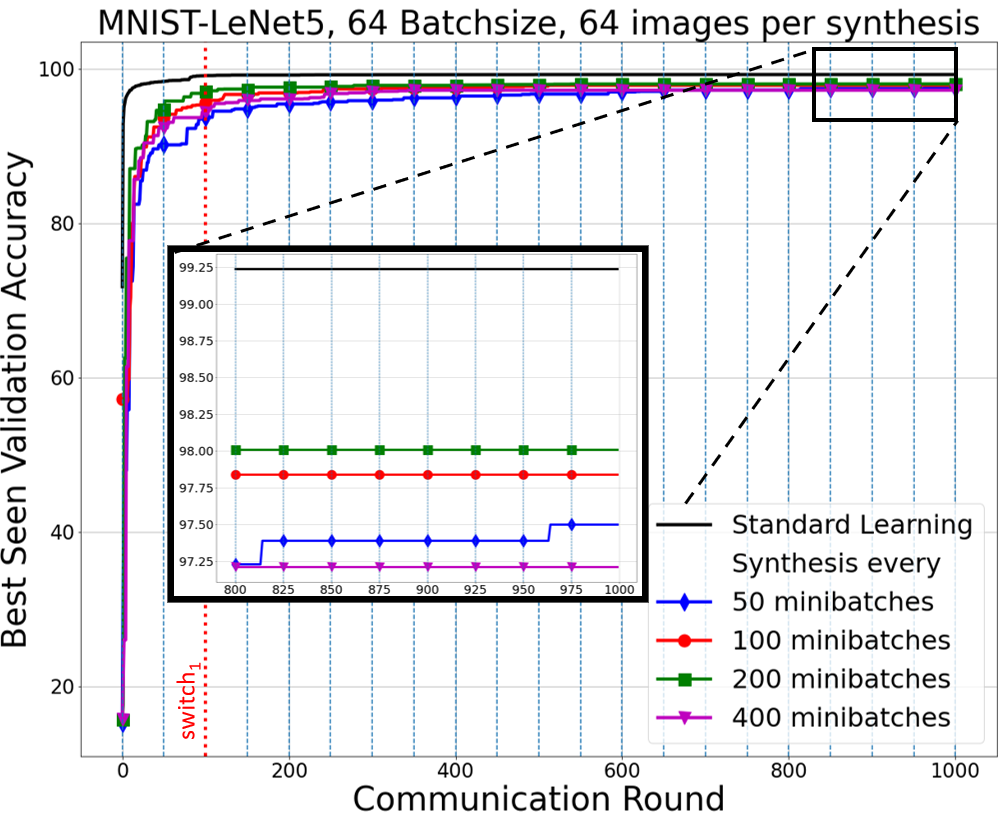}
         \caption{Variation in frequency of synthesis.}
         \label{fig:MNIST_single_freq}
     \end{subfigure}
        \caption{Single Device Performance of synthetic images for the MNIST dataset trained on a LeNet5 architecture.}
        \label{fig:MNIST_single_device}
\end{figure}

\begin{figure}[!h]
     \centering
     \begin{subfigure}[]{0.95\columnwidth}
         \centering
         \includegraphics[width=\textwidth]{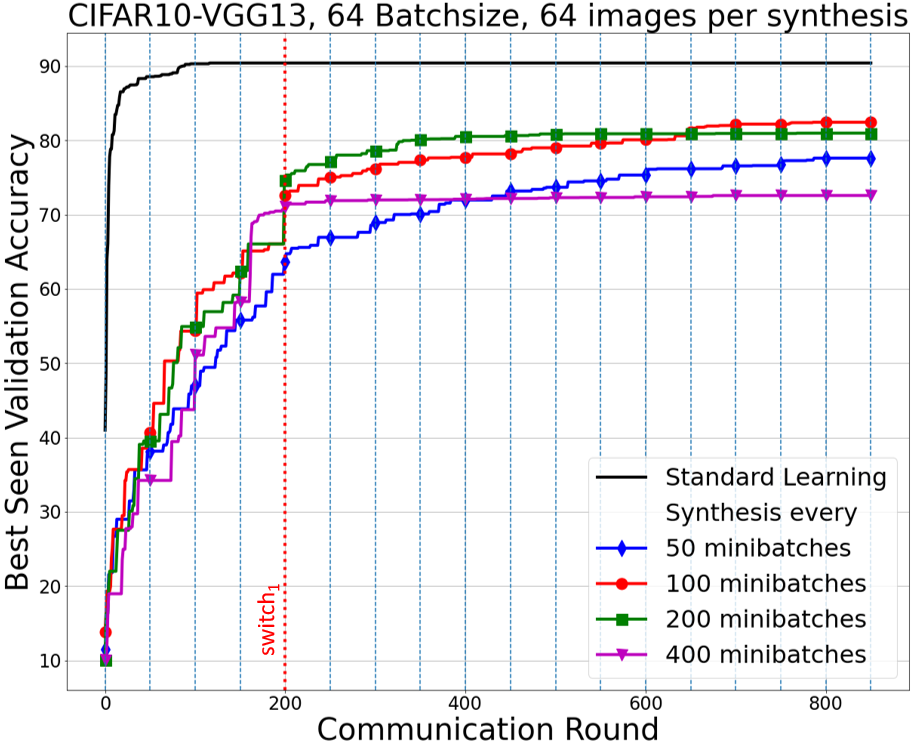}
         \caption{Variation in frequency of synthesis.}
         \label{fig:CIFAR_single_freq}
     \end{subfigure}
        \hfill
     \begin{subfigure}[]{0.95\columnwidth}
         \centering
         \includegraphics[width=\textwidth]{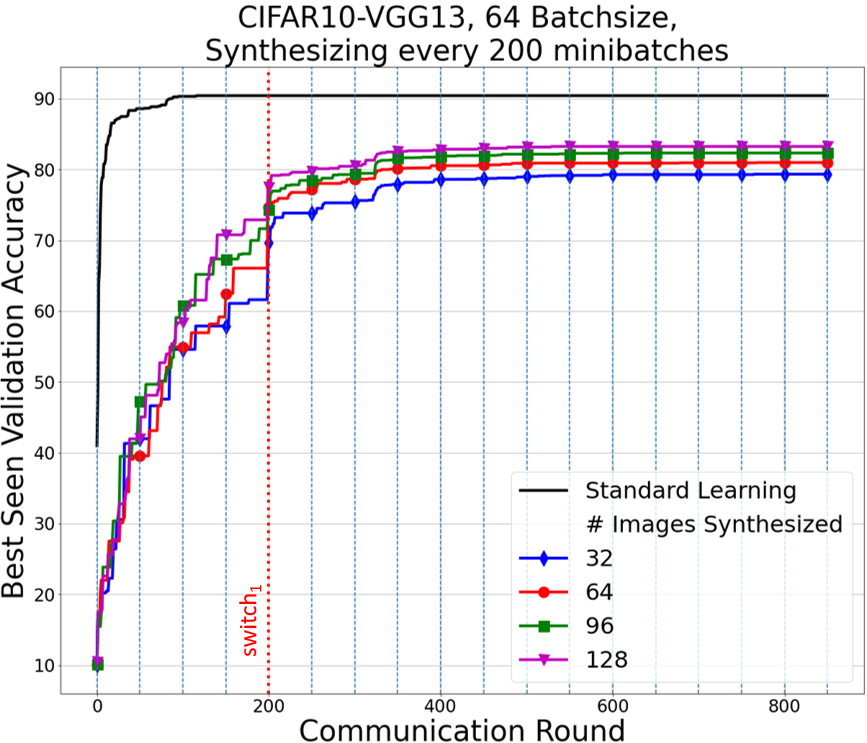}
         \caption{Variation of number of synthetic images.}
         \label{fig:CIFAR_single_nimgs}
     \end{subfigure}   
        \caption{Single Device Performance of synthetic images for the CIFAR10 dataset trained on a VGG13 architecture.}
        \label{fig:CIFAR_single_device_0}
\end{figure}

We now show the learning curves to show how validation accuracy varies with communication rounds. Figures \ref{fig:MNIST_single_device} and \ref{fig:CIFAR_single_device_0} show the performance of TOFU on MNIST and CIFAR-10 on a single device platforms, for different \textit{Synfreq} and \textit{Nimgs}. We set $switch_1$ to $100$ communication rounds in MNIST and $200$ communication rounds in CIFAR-10. We do not have phase 3 for single device learning.

\subsubsection{Federated Performance on MNIST}
\label{appendix:federated_mnist}
For the MNIST dataset, the data was distributed among 10 clients. We use a participation rate of 1, and all clients participate in all rounds. A summary of our results is tabulated in Table \ref{Table:Federated_MNIST}. Note that the number of communication rounds denoted are the ones to achieve the maximum accuracy. 
\begin{table}[!h]
\centering
\def\arraystretch{1.1}%

\setlength{\tabcolsep}{3.4pt}
\begin{tabular}{|cccccc|}
\hline
\multicolumn{6}{|c|}{MNIST, LeNet, Number of Clients= 10, IID Distribution} \\ \hline
\multicolumn{1}{|c|}{} & \multicolumn{3}{c|}{Using Only Synthetic Data} & \multicolumn{2}{c|}{} \\ \cline{1-4}
\multicolumn{1}{|c|}{} & \multicolumn{1}{c|}{} & \multicolumn{1}{c|}{} & \multicolumn{1}{c|}{} & \multicolumn{2}{c|}{\multirow{-2}{*}{\begin{tabular}[c]{@{}c@{}}+ 3 Rounds of\\  FedAvg\end{tabular}}} \\ \cline{5-6} 
\multicolumn{1}{|c|}{\multirow{-2}{*}{}} & \multicolumn{1}{c|}{\multirow{-2}{*}{\begin{tabular}[c]{@{}c@{}}Max \\Acc. (\%)\end{tabular}}} & \multicolumn{1}{c|}{\multirow{-2}{*}{\begin{tabular}[c]{@{}c@{}}Comm.\\ Rnds.\end{tabular}}} & \multicolumn{1}{c|}{\multirow{-2}{*}{\begin{tabular}[c]{@{}c@{}}Comm.\\ Eff.\end{tabular}}} &  &  \\ \cline{1-4}
\multicolumn{1}{|c|}{\cellcolor[HTML]{C0C0C0}FEDAVG} & \multicolumn{1}{c|}{\cellcolor[HTML]{C0C0C0}98.91} & \multicolumn{1}{c|}{\cellcolor[HTML]{C0C0C0}456} & \multicolumn{1}{c|}{\cellcolor[HTML]{C0C0C0}1.0$\times$} & \multirow{-2}{*}{\begin{tabular}[c]{@{}c@{}}Max \\Acc. (\%)\end{tabular}} & \multirow{-2}{*}{\begin{tabular}[c]{@{}c@{}}Comm.\\ Eff.\end{tabular}} \\ \hline
\multicolumn{1}{|c|}{Nimgs} & \multicolumn{5}{c|}{Varying Nimgs @ Synfreq = 1 local epoch} \\ \hline
\multicolumn{1}{|c|}{32} & 95.39 & 129 & \multicolumn{1}{c|}{6.9$\times$} & 98.06 & 6.6$\times$ \\
\multicolumn{1}{|c|}{64} & 96.06 & 104 & \multicolumn{1}{c|}{4.2$\times$} & 98.14 & 4.1$\times$ \\
\multicolumn{1}{|c|}{96} & 96.77 & 83 & \multicolumn{1}{c|}{3.6$\times$} & 98.01 & 3.5$\times$ \\
\multicolumn{1}{|c|}{128} & 95.23 & 53 & \multicolumn{1}{c|}{4.2$\times$} & 98.07 & 4.1$\times$ \\ \hline
\multicolumn{1}{|c|}{Synfreq} & \multicolumn{5}{c|}{Varying Synfreq @ Nimgs =64} \\ \hline
\multicolumn{1}{|c|}{25} & 92.00 & 130 & \multicolumn{1}{c|}{3.4$\times$} & 98.29 & 3.3$\times$ \\
\multicolumn{1}{|c|}{50} & 91.71 & 56 & \multicolumn{1}{c|}{7.9$\times$} & 98.19 & 7.5$\times$ \\
\multicolumn{1}{|c|}{1 ep} & 96.06 & 104 & \multicolumn{1}{c|}{4.2$\times$} & 98.14 & 4.1$\times$ \\
\multicolumn{1}{|c|}{2 eps} & 95.52 & 142 & \multicolumn{1}{c|}{3.1$\times$} & 97.91 & 3.1$\times$ \\ \hline
\end{tabular}
\caption{Table showing Federated platform accuracies on MNIST. The baseline of FedAvg is shown in grey. }
\label{Table:Federated_MNIST}
\end{table}
\begin{table}[ht]

\label{Table:IsoFederated_MNIST}
\def\arraystretch{1.1}%
\setlength{\tabcolsep}{5.2pt}

\begin{tabular}{|cccccc|}
\hline
\multicolumn{6}{|c|}{\begin{tabular}[c]{@{}c@{}}MNIST, LeNet5, Number of Clients =10\\ IID Distribution\end{tabular}} \\ \hline
\multicolumn{1}{|c|}{} & \multicolumn{2}{c|}{Accuracy : 95\%} & \multicolumn{3}{c|}{Accuracy : 98\%} \\ \cline{2-6} 
\multicolumn{1}{|c|}{} & \multicolumn{1}{c|}{} & \multicolumn{1}{c|}{} & \multicolumn{1}{c|}{} & \multicolumn{1}{c|}{} &  \\
\multicolumn{1}{|c|}{\multirow{-3}{*}{}} & \multicolumn{1}{c|}{\multirow{-2}{*}{\begin{tabular}[c]{@{}c@{}}Comm\\ Rnds\end{tabular}}} & \multicolumn{1}{c|}{\multirow{-2}{*}{\begin{tabular}[c]{@{}c@{}}Comm.\\ Eff.\end{tabular}}} & \multicolumn{1}{c|}{\multirow{-2}{*}{\begin{tabular}[c]{@{}c@{}}Syn.\\ Rnds\end{tabular}}} & \multicolumn{1}{c|}{\multirow{-2}{*}{\begin{tabular}[c]{@{}c@{}}FedAvg\\ Rnds\end{tabular}}} & \multirow{-2}{*}{\begin{tabular}[c]{@{}c@{}}Comm.\\ Eff.\end{tabular}} \\ \hline
\rowcolor[HTML]{C0C0C0} 
\multicolumn{1}{|c|}{\cellcolor[HTML]{C0C0C0}FEDAVG} & 12 & \multicolumn{1}{c|}{\cellcolor[HTML]{C0C0C0}1.0$\times$} & - & 48 & 1.0$\times$ \\ \hline
\multicolumn{1}{|c|}{Nimgs} & \multicolumn{5}{c|}{Varying Nimgs @ synfreq = 1 local epoch} \\ \hline
\multicolumn{1}{|c|}{32} & 121 & \multicolumn{1}{c|}{0.2$\times$} & 129 & 3 & 0.7$\times$ \\ 
 \multicolumn{1}{|c|}{64} & 55 & \multicolumn{1}{c|}{0.2$\times$} & 104 & 3 & 0.8$\times$ \\ 
\multicolumn{1}{|c|}{96} & 60 & \multicolumn{1}{c|}{0.1$\times$} & 526 & 3 & 0.5$\times$ \\ 
\multicolumn{1}{|c|}{128} & 53 & \multicolumn{1}{c|}{0.1$\times$} & 439 & 3 & 0.4$\times$ \\ \hline
\end{tabular}
\caption{Federated results for Isoaccuracy on MNIST. The baseline of FedAvg is shown in grey. `Syn. Rnds' denotes number of communication rounds using synthetic data while `FedAvg Rnds' denotes the number of rounds for full weight update exchange.}
\label{Table:IsoFederated_MNIST}
\end{table}

We first vary the size of the synthetic dataset (Number of images or Nimgs) from 32 Images to 128 Images. We find that although a larger number of images reaches a higher accuracy faster, it loses out on efficiency. We also show the accuracy results by varying the frequency of synthesis by keeping the number of images constant. We empirically found that synthesizing and communication after 1 local epoch (97 minibatches) achieved the highest accuracy, while synthesizing and communicating after every 50 minibatches provided the most efficiency. We were able to achieve less that a 1\% drop in accuracy after exchanging only 3 full weight updates.

For a fairer comparison, we also report results on iso-accuracy, since the baseline FedAvg reaches higher accuracies, which will naturally take more communication rounds and skew efficiency in our favor. Table \ref{Table:IsoFederated_MNIST} shows the number of communication rounds needed to achieve iso-accuracy with the baseline (FedAvg). For 95\%, only the synthetic images are sufficient, for 98\%, additional full weight updates are needed.

\begin{figure}[h]
     \centering
     \begin{subfigure}[!t]{0.89\columnwidth}
         \centering
\includegraphics[width=\textwidth]{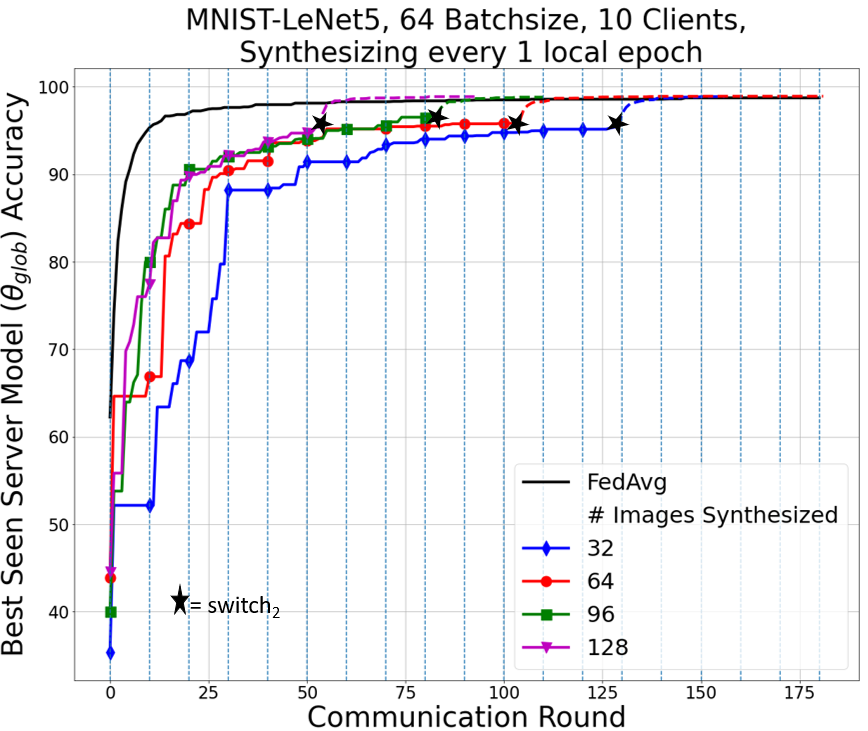}
         \caption{Variation of number of synthetic images.}
         \label{fig:MNIST_fed_nimgs}
     \end{subfigure}
     \hfill
     \begin{subfigure}[!t]{0.89\columnwidth}
         \centering
         \includegraphics[width=\textwidth]{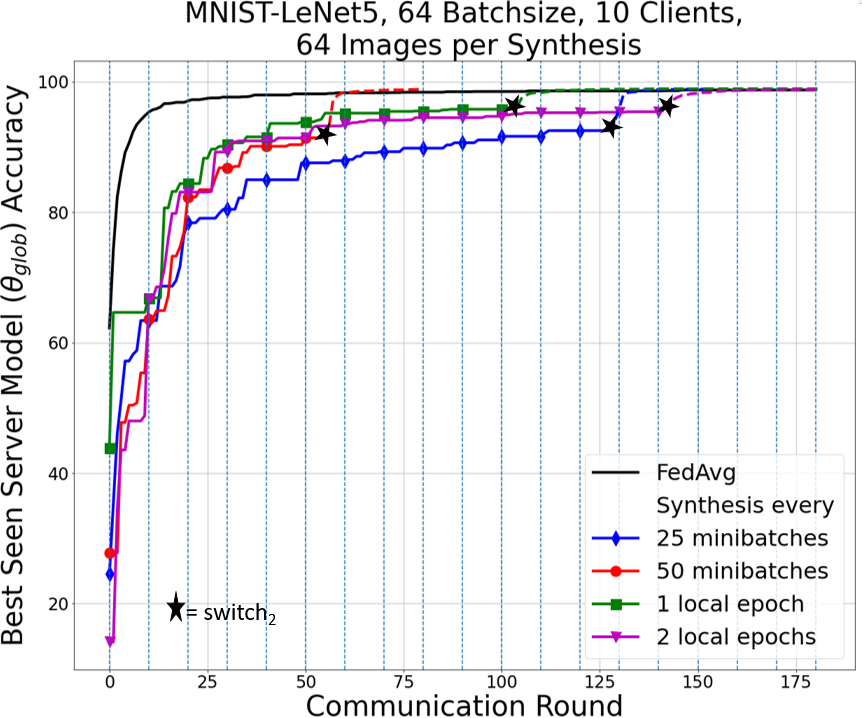}
         \caption{Variation in frequency of synthesis.}
         \label{fig:MNIST_fed_freq}
     \end{subfigure}
        \caption{Federated performance of TOFU on the MNIST dataset, distributed in an IID fashion amongst 10 clients. We finetuned $switch_{2}$ for efficiency and is denoted on the graphs by a star. We found that in the case of MNIST, $switch_{1}$ was not helpful.}
        \label{fig:MNIST_fed_platform}
\end{figure}

Figure \ref{fig:MNIST_fed_platform} shows the learning curves. We found that on the MNIST dataset, $switch_{1}$ did not provide much of a benefit. $switch_2$ was manually fine tuned in order to achieve the highest communication efficiency. The dashed lines on the graph represent the additional full weight updates. 

\begin{table}[!b]

\label{Table:IsoFederated_CIFAR}
\def\arraystretch{1.1}%

\begin{tabular}{|cccccc|}
\hline
\multicolumn{6}{|c|}{\begin{tabular}[c]{@{}c@{}}CIFAR10, VGG13, Number of Clients =5, IID Distribution\end{tabular}} \\ \hline
\multicolumn{1}{|c|}{} & \multicolumn{2}{c|}{Accuracy : 72\%} & \multicolumn{3}{c|}{Accuracy : 87\%} \\ \cline{2-6} 
\multicolumn{1}{|c|}{} & \multicolumn{1}{c|}{} & \multicolumn{1}{c|}{} & \multicolumn{1}{c|}{} & \multicolumn{1}{c|}{} &  \\
\multicolumn{1}{|c|}{\multirow{-3}{*}{}} & \multicolumn{1}{c|}{\multirow{-2}{*}{\begin{tabular}[c]{@{}c@{}}Comm\\ Rnds\end{tabular}}} & \multicolumn{1}{c|}{\multirow{-2}{*}{\begin{tabular}[c]{@{}c@{}}Comm.\\ Eff.\end{tabular}}} & \multicolumn{1}{c|}{\multirow{-2}{*}{\begin{tabular}[c]{@{}c@{}}Syn.\\ Rnds\end{tabular}}} & \multicolumn{1}{c|}{\multirow{-2}{*}{\begin{tabular}[c]{@{}c@{}}FedAvg\\ Rnds\end{tabular}}} & \multirow{-2}{*}{\begin{tabular}[c]{@{}c@{}}Comm.\\ Eff.\end{tabular}} \\ \hline
\rowcolor[HTML]{C0C0C0} 
\multicolumn{1}{|c|}{\cellcolor[HTML]{C0C0C0}FEDAVG} & 25 & \multicolumn{1}{c|}{\cellcolor[HTML]{C0C0C0}1.0$\times$} & - & 82 & 1.0$\times$ \\ \hline
\multicolumn{1}{|c|}{Nimgs} & \multicolumn{5}{c|}{Varying Nimgs @ synfreq = 1 local epoch} \\ \hline
\multicolumn{1}{|c|}{64} & 477 & \multicolumn{1}{c|}{2.5$\times$} & 477 & 6 & 5.1$\times$ \\ 
\multicolumn{1}{|c|}{96} & 461 & \multicolumn{1}{c|}{1.7$\times$} & 526 & 6 & 3.6$\times$ \\ 
\multicolumn{1}{|c|}{128} & 403 & \multicolumn{1}{c|}{1.5$\times$} & 439 & 6 & 3.4$\times$ \\ \hline
\end{tabular}
\caption{Federated results for Iso-accuracy on CIFAR-10. The baseline of FedAvg is shown in grey. We show results for reaching 72\% and 87\% accuracy. `Syn. Rnds' denotes the number of communication rounds using synthetic data while 'FedAvg Rnds' denotes the number of rounds for full weight update exchange.}
\label{Table:IsoFederated_CIFAR}
\end{table}

\subsubsection{Federated Performance on CIFAR-10}
\label{appendix:federated_cifar}
For CIFAR-10, the data was distributed among 5 clients. We use a participation rate of 1, and all clients participate in all rounds. Our results are summarized in Table \ref{Table:Federated_CIFAR}.

\begin{table*}[!ht]
\renewcommand{\arraystretch}{1.2}
\begin{tabular}{|ccccccc|}

\hline
\multicolumn{7}{|c|}{CIFAR-10, VGG13, Number of Clients= 5, IID Distribution} \\ \hline
\multicolumn{1}{|c|}{} & \multicolumn{3}{c|}{Using Only Synthetic Data} & \multicolumn{3}{c|}{\begin{tabular}[c]{@{}c@{}}Additional Comm Rounds of\\  Full Gradient Exchange\end{tabular}} \\ \hline
\multicolumn{1}{|c|}{} & \multicolumn{1}{c|}{} & \multicolumn{1}{c|}{} & \multicolumn{1}{c|}{} & \multicolumn{1}{c|}{} & \multicolumn{1}{c|}{} &  \\
\multicolumn{1}{|c|}{\multirow{-2}{*}{}} & \multicolumn{1}{c|}{\multirow{-2}{*}{\begin{tabular}[c]{@{}c@{}}Max Accuracy \\ (\%)\end{tabular}}} & \multicolumn{1}{c|}{\multirow{-2}{*}{\begin{tabular}[c]{@{}c@{}}Comm.\\ Rounds\end{tabular}}} & \multicolumn{1}{c|}{\multirow{-2}{*}{\begin{tabular}[c]{@{}c@{}}Comm.\\ Efficiency\end{tabular}}} & \multicolumn{1}{c|}{} & \multicolumn{1}{c|}{} &  \\ \cline{1-4}
\multicolumn{1}{|c|}{\cellcolor[HTML]{C0C0C0}FEDAVG} & \multicolumn{1}{c|}{\cellcolor[HTML]{C0C0C0}88.73} & \multicolumn{1}{c|}{\cellcolor[HTML]{C0C0C0}166} & \multicolumn{1}{c|}{\cellcolor[HTML]{C0C0C0}1.0$\times$} & \multicolumn{1}{c|}{\multirow{-3}{*}{\begin{tabular}[c]{@{}c@{}}+ 5 Rounds\\ Accuracy \% \\ (Comm. Efficiency)\end{tabular}}} & \multicolumn{1}{c|}{\multirow{-3}{*}{\begin{tabular}[c]{@{}c@{}}+ 10 Rounds\\ Accuracy \%\\ (Comm. Efficiency)\end{tabular}}} & \multirow{-3}{*}{\begin{tabular}[c]{@{}c@{}}+15 Rounds\\ Accuracy \%\\ (Comm. Efficiency)\end{tabular}} \\ \hline
\multicolumn{1}{|c|}{Nimgs} & \multicolumn{6}{c|}{Varying Nimgs @ Synfreq = 1 local epoch} \\ \hline
\multicolumn{1}{|c|}{32} & 70.81 & 593 & \multicolumn{1}{c|}{26.7$\times$} & \begin{tabular}[c]{@{}c@{}}86.12  (14.8$\times$)\end{tabular} & \begin{tabular}[c]{@{}c@{}}87.50 (10.2$\times$)\end{tabular} & \begin{tabular}[c]{@{}c@{}}88.31 (7.8$\times$)\end{tabular} \\ 
\multicolumn{1}{|c|}{64} & 72.31 & 477 & \multicolumn{1}{c|}{16.6$\times$} & \begin{tabular}[c]{@{}c@{}}86.80 (11.1$\times$)\end{tabular} & \begin{tabular}[c]{@{}c@{}}87.92 (8.3$\times$)\end{tabular} & \begin{tabular}[c]{@{}c@{}}88.34 (6.6$\times$)\end{tabular} \\ 
\multicolumn{1}{|c|}{96} & 74.06 & 526 & \multicolumn{1}{c|}{10.0$\times$} & \begin{tabular}[c]{@{}c@{}}86.06 (7.7$\times$)\end{tabular} & \begin{tabular}[c]{@{}c@{}}87.77 (6.3$\times$)\end{tabular} & \begin{tabular}[c]{@{}c@{}}88.48 (5.3$\times$)\end{tabular} \\ 
\multicolumn{1}{|c|}{128} & 74.03 & 439 & \multicolumn{1}{c|}{9.0$\times$} & \begin{tabular}[c]{@{}c@{}}86.43 (7.0$\times$)\end{tabular} & \begin{tabular}[c]{@{}c@{}}87.90 (5.8$\times$)\end{tabular} & \begin{tabular}[c]{@{}c@{}}88.34 (4.9$\times$)\end{tabular} \\ \hline
\multicolumn{1}{|c|}{Synfreq} & \multicolumn{6}{c|}{Varying Synfreq @ Nimgs =64} \\ \hline
\multicolumn{1}{|c|}{50} & 63.46 & 414 & \multicolumn{1}{c|}{19.1$\times$} & \begin{tabular}[c]{@{}c@{}}86.72 (12.1$\times$)\end{tabular} & \begin{tabular}[c]{@{}c@{}}87.86 (8.9$\times$)\end{tabular} & \begin{tabular}[c]{@{}c@{}}88.36 (7.0$\times$)\end{tabular} \\
\multicolumn{1}{|c|}{100} & 68.21 & 511 & \multicolumn{1}{c|}{15.5$\times$} & \begin{tabular}[c]{@{}c@{}}86.48 (10.6$\times$)\end{tabular} & \begin{tabular}[c]{@{}c@{}}88.05 (8.0$\times$)\end{tabular} & \begin{tabular}[c]{@{}c@{}}88.54 (6.5$\times$)\end{tabular} \\ 
\multicolumn{1}{|c|}{1 epoch} & 72.31 & 477 & \multicolumn{1}{c|}{16.6$\times$} & \begin{tabular}[c]{@{}c@{}}86.80 (11.1$\times$)\end{tabular} & \begin{tabular}[c]{@{}c@{}}87.92 (8.3$\times$)\end{tabular} & \begin{tabular}[c]{@{}c@{}}88.34 (6.6$\times$)\end{tabular} \\ \hline
\end{tabular}
\caption{Table showing Federated platform accuracies on CIFAR-10. The baseline of FedAvg is shown in grey. }

\label{Table:Federated_CIFAR}
\end{table*}

\begin{figure*}[!h]
     \centering
     \begin{subfigure}[t]{0.9\columnwidth}
         \centering
\includegraphics[width=\textwidth]{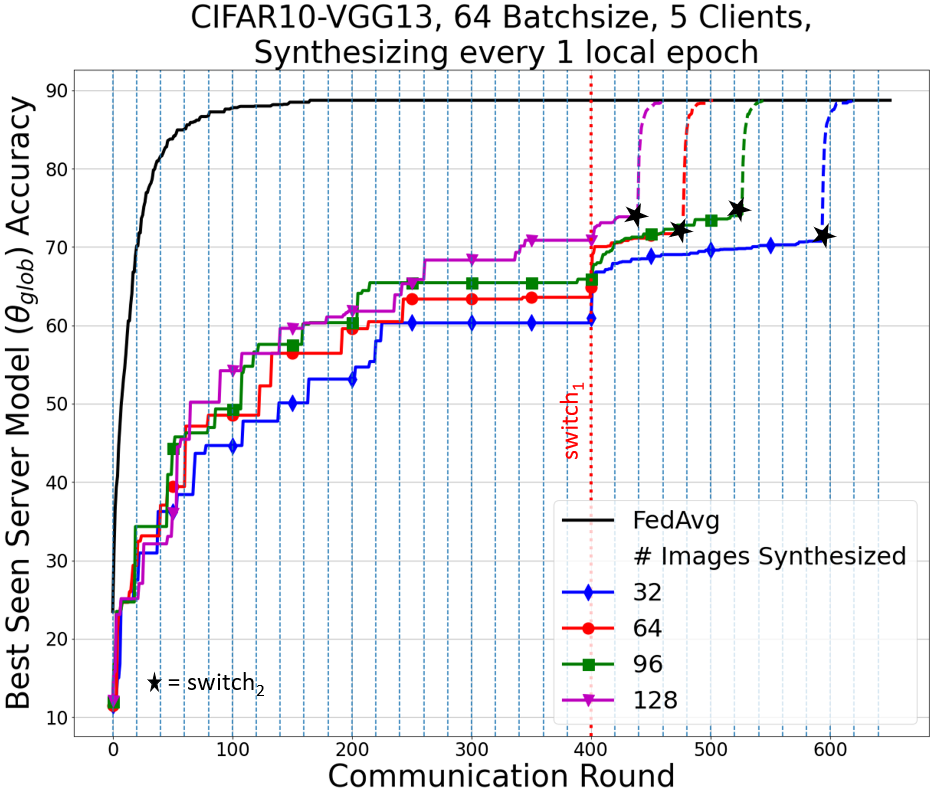}
         \caption{Variation of number of synthetic images.}
         \label{fig:CIFAR_fed_nimgs}
     \end{subfigure}
     \hfill
     \begin{subfigure}[t]{0.9\columnwidth}
         \centering
         \includegraphics[width=\textwidth]{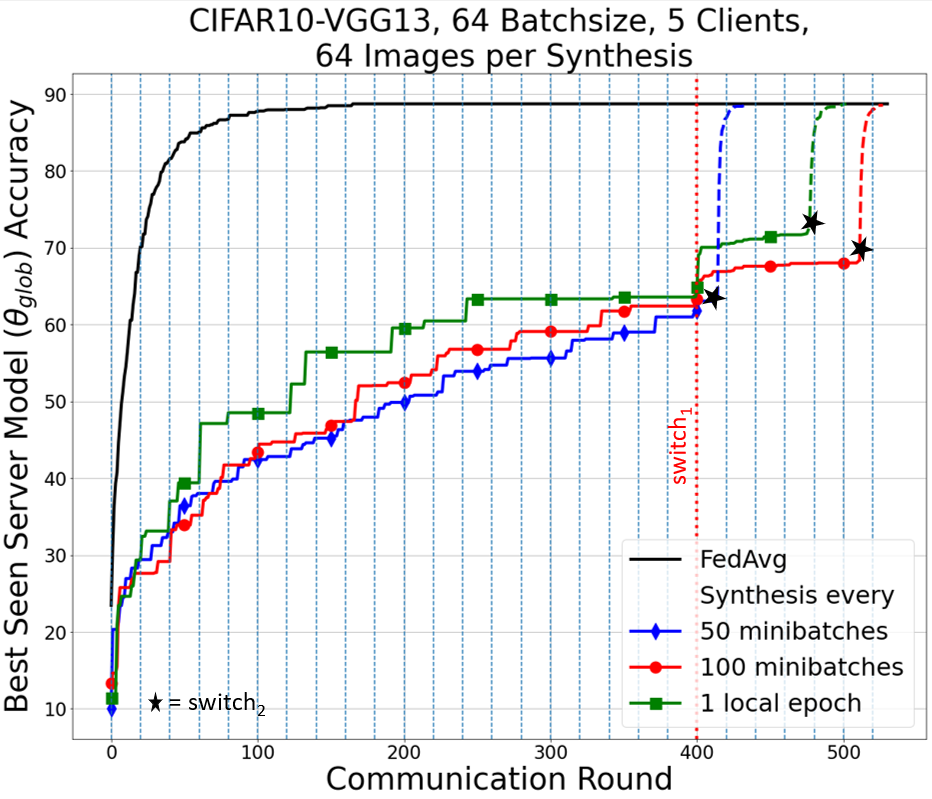}
         \caption{Variation in frequency of synthesis.}
         \label{fig:CIFAR_fed_freq}
     \end{subfigure}
        \caption{Federated Performance of TOFU on the CIFAR-10 dataset, distributed in an IID fashion amongst 5 clients. We set $switch_{1}$ to 400 for all runs. The $switch_{2}$ was finetuned to provide optimal efficiency and is denoted on the graphs by a star.}
        \label{fig:CIFAR_fed_platform}
\end{figure*}

Similar to MNIST, we first vary the number of images (size of the synthetic dataset) from 32 images to 128 images. We achieve a higher accuracy with larger number of images, but the communication efficiency reduces. We then fix the number of images to be 64 and vary the frequency of synthesis. We found that synthesizing after every 50 minibatches (roughly 1/3rd of an epoch) produces the most efficient results. We also show that we can achieve less than a 3\% drop with 5 additional weight updates and less than a 1\% drop with 15 additional weight updates. Table \ref{Table:IsoFederated_CIFAR} shows the number of communication rounds needed to achieve iso-accuracy using TOFU, with synthesis after every local epoch. To achieve 72\%, only the synthetic images were sufficient, for 87\% 6 additional full weight update rounds were needed. The corresponding communication efficiency numbers are shown.

Figure \ref{fig:CIFAR_fed_platform} shows the corresponding learning curves. We found that on the CIFAR-10 dataset, $switch_{1}$ was set to 400 communication rounds. $switch_2$ was manually fine tuned in order to achieve the highest communication efficiency and is denoted by a star for each curve. The dashed lines on the graph represent the additional full weight updates.


\end{document}